\documentclass{article}

\usepackage{amsmath,amsfonts}

\DeclareMathOperator*{\argmin}{arg\,min}
\usepackage{algorithmic}
\usepackage{algorithm}
\usepackage{array}
\usepackage{textcomp}
\usepackage{stfloats}
\usepackage{verbatim}
\usepackage{cite}
\usepackage{multirow}

\usepackage{PRIMEarxiv}
\usepackage[utf8]{inputenc} 
\usepackage[T1]{fontenc}    
\usepackage{hyperref}       
\usepackage{url}            
\usepackage{booktabs}       
\usepackage{nicefrac}       
\usepackage{microtype}      
\usepackage{lipsum}
\usepackage{fancyhdr}       
\usepackage{graphicx}       
\graphicspath{{media/}}     

\title{Meta-Simulation for the Automated Design of Synthetic Overhead Imagery
}

\author{
  Handi Yu, Simiao Ren, Leslie M. Collins \\
  Department of Electrical and Computer Engineering \\
  Duke University \\
  Durham, NC 27705, USA\\
  \texttt{hdyu1030@gmail.com, simiao.ren@duke.edu, leslie.collins@duke.edu} \\
   \And
  Jordan M. Malof \\
  Department of Computer Science \\
  University of Montana \\
  Missoula, MT 59812, USA\\
  \texttt{jmmalo03@gmail.com} \\
}

\begin{document}
\maketitle

\begin{abstract}
The use of synthetic (or simulated) data for training machine learning models has grown rapidly in recent years. Synthetic data can often be generated much faster and more cheaply than its real-world counterpart. One challenge of using synthetic imagery however is scene design: e.g., the choice of content and its features and spatial arrangement. To be effective, this design must not only be realistic, but appropriate for the target domain, which (by assumption) is unlabeled. In this work, we propose an approach to automatically choose the design of synthetic imagery based upon unlabeled real-world imagery. Our approach, termed Neural-Adjoint Meta-Simulation (NAMS), builds upon the seminal recent meta-simulation approaches. In contrast to the current state-of-the-art methods, our approach can be pre-trained once offline, and then provides fast design inference for new target imagery. Using both synthetic and real-world problems, we show that NAMS infers synthetic designs that match both the in-domain and out-of-domain target imagery, and that training segmentation models with NAMS-designed imagery yields superior results compared to naïve randomized designs and state-of-the-art meta-simulation methods.
\end{abstract}


\section{Introduction} 
\label{sec:introduction}

A well-known challenge of modern high-capacity recognition models, such as deep neural networks (DNNs), is their need for large quantities of training data. This is a major problem for tasks involving overhead (e.g., satellite) imagery due to the costs to purchase and label the imagery. Furthermore, there is tremendous variability in real-world overhead imagery (e.g., due to geography, weather, time-of-day, imaging hardware, and more), making it costly to capture in a dataset. Despite their tremendous benefits for the research community, recent benchmark datasets of overhead imagery (e.g., DSTL \cite{Iglovikov2018}, DeepGlobe \cite{Demir2018}, and Inria \cite{Maggiori2017}) encompass just a few geographic locations and environmental conditions. As a result, these datasets still capture a small fraction of real-world visual variability, and DNNs trained on these datasets have been found to perform unpredictably, and usually poorly, on new collections of overhead imagery \cite{Maggiori2017,Wang2017,Huang2018,Kong2020}. This limitation greatly undermines their value to researchers and real-world users. 


In this work we explore the use of \textit{synthetic} overhead imagery to address this problem.  Synthetic imagery is captured using a virtual camera operating in a virtual world, where the designer can control the scene content and environmental conditions. By systematically varying the camera properties, scene content, and environmental conditions it is possible to rapidly capture large quantities of imagery. Furthermore, because we have all information about the camera and the scene, we can also automatically generate a variety of high-quality ground truth annotations, such as full pixel-wise segmentation of scene content and objects, or image depth maps.  Such annotations are often expensive or impracticable to collect.   For these reasons, synthetic imagery potentially offers a fast and cost-effective means to collect large quantities of diverse training imagery for DNNs. Due to these potential benefits, interest in synthetic imagery has grown rapidly in recent years, and a large number of studies have demonstrated that it is beneficial for training DNNs on a variety of tasks \cite{Kong2020,Ward2018,Shermeyer2021,Xu,Hu2021,Han2017,Ros2016,Richter2016,Tobin2017,Zhang2017,Shafaei2016,Qureshi2008,Taylor2007,Tremblay2018,Sankaranarayanan2018}, including those involving overhead imagery \cite{Kong2020,Ward2018,Shermeyer2021,Xu,Hu2021,Han2017}.  



\begin{figure}
    \centering
    \includegraphics[height=5.6cm]{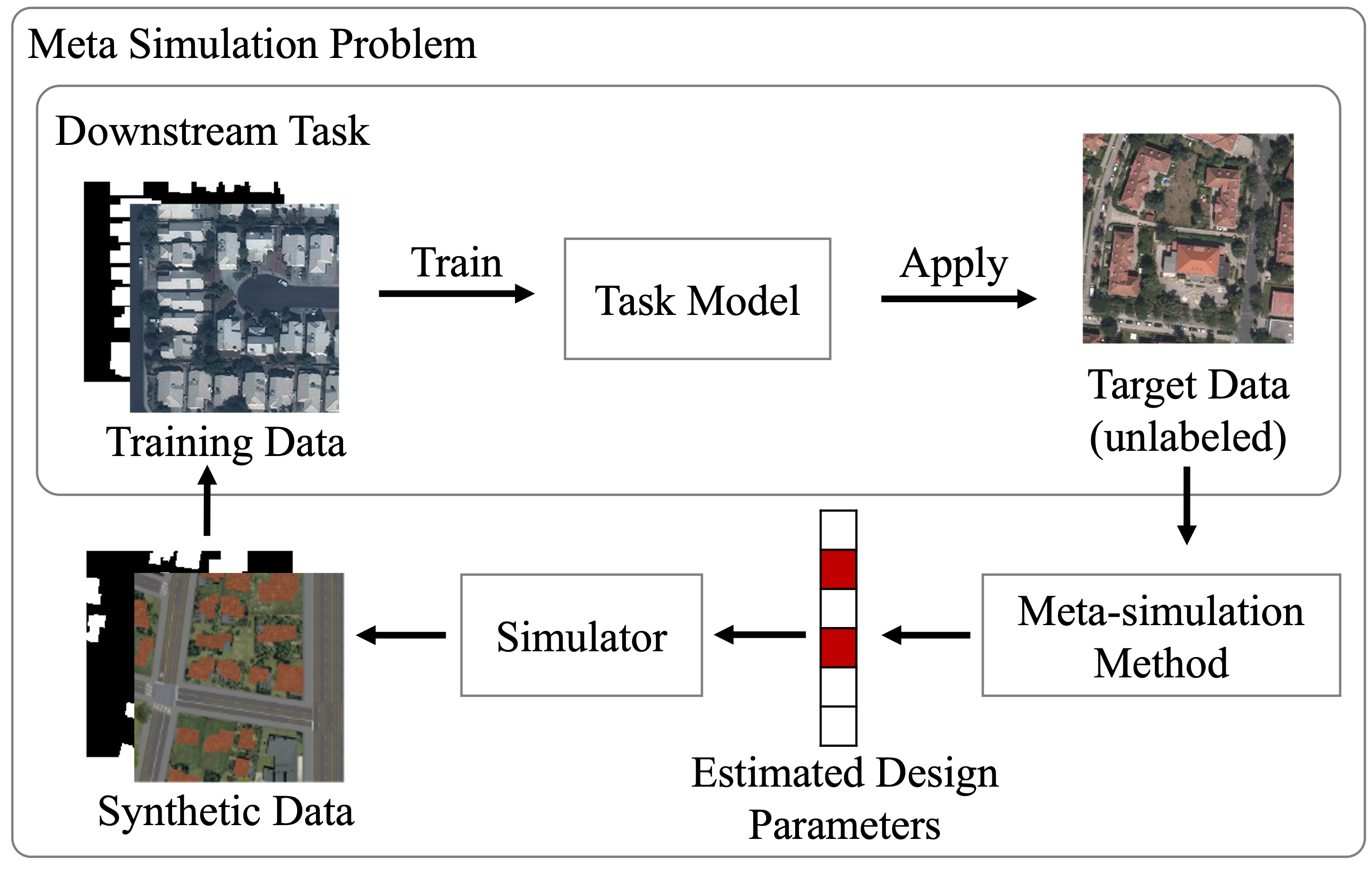}
    \caption{Overview of the meta-simulation problem formulation: the goal is to estimate the target design parameters and generate realistic synthetic images in the target domain for augmenting the training of a downstream task model.}
    \label{fig:overview}
\end{figure}

One major obstacle to the success of synthetic imagery is the \textit{sim-to-real gap} \cite{Kar2019a,Devaranjan2020}, which refers to systematic visual dissimilarities between the synthetic and real-world imagery.  A large body of research has been focused on overcoming this gap \cite{Tremblay2018,Sankaranarayanan2018,Kar2019a,Devaranjan2020,hoffman_2018,huang2018multimodal}. The majority of this work has focused on overcoming so-called \textit{appearance} gaps, which refers to gaps caused primarily by limitations in the graphics engine that affect low-level imagery features, such as shading and texture \cite{hoffman_2018, huang2018multimodal}. Relatively less attention has been given to overcoming the \textit{content} gap, which refers to differences in scene content between the virtual world and real-world scenes - these gaps then result in significant visual differences between the imagery.  Content gaps can include differences in the types of objects in the scene, their material composition (which often affects object texture and color), and their spatial arrangement \cite{Kar2019a}. Content gaps can be overcome through careful manual design, however this is costly and time-consuming \cite{Kong2020}.  Furthermore, the design process must generally be repeated for each new context, or visual domain, one wishes to perform well upon.  This has led to the recent development of \textit{meta-simulation} models which aim to (semi-) automate the design of the virtual world, mitigating one of the major limitations of synthetic imagery  \cite{Kar2019a,Devaranjan2020,Kulkarni2015Deep,Du2021Auto-Tuned,Mansinghka2013Approximate,Kulkarni2015Picture:,Yildirim2015,Louppe2019Adversarial,Ganin2018Synthesizing,Mellor2019Unsupervised,Behl2020AutoSimulate:,Ruiz2019LEARNING}.  


\subsection{Meta-simulation} 
\label{subsec:introduction_meta_simulation}

Meta-simulation methods attempt to optimize the design parameters of the simulator (which primarily influence the virtual scene content) in order to generate synthetic images that look similar to those in a set of \textit{unlabeled} target imagery, as illustrated in Fig. \ref{fig:overview}.  We can then use the synthetic imagery for training a task model to perform well on the target unlabeled imagery.  Solving the meta-simulation problem is difficult because we must simultaneously infer the types of content in the scene, their spatial arrangement and quantity, and their material  composition  - a vast search space.  The optimization process is made more difficult because simulators are typically complex and non-differentiable black-box functions, and may also have non-ordinal categorical input parameters. 



Despite these challenges, a variety of meta-simulation methods have been proposed in recent years \cite{Kar2019a,Devaranjan2020,Kulkarni2015Deep,Du2021Auto-Tuned,Mansinghka2013Approximate,Kulkarni2015Picture:,Yildirim2015,Louppe2019Adversarial,Ganin2018Synthesizing,Mellor2019Unsupervised,Behl2020AutoSimulate:,Ruiz2019LEARNING}.  In these seminal works, the authors demonstrated that meta-simulation can be used to infer appropriate design parameters, as well as improve the accuracy of recognition models trained on the resulting synthetic data.  However, most of this existing work has focused upon natural imagery, which exhibits some important differences with respect to overhead imagery.  First, there exist simulation engines for natural imagery that are computationally fast, and produce relatively realistic synthetic imagery (i.e., a low sim-to-real visual gap). By contrast, publicly-accessible simulators for overhead imagery are substantially slower, and exhibit a relatively large sim-to-real visual gap (see Sec. \ref{sec:related_work_syn}).  Another important distinction is that natural imagery is relatively easy to collect, and it is therefore often abundant. By contrast, overhead imagery is costly to collect, and is often limited in quantity for novel target domains. .    

Due to these differences, many existing meta-simulation methods are not directly applicable to overhead imagery (see Sec. \ref{subsec:related_work_MS}) and there has been little work investigating meta-simulation specifically for overhead imagery. The authors in \cite{Kar2019a} and \cite{Devaranjan2020} demonstrated the effectiveness of several meta-simulation methods on overhead imagery - one of which, termed Meta-Sim2 (MS2) - we also investigate here.  However these works only tested on synthetic target imagery (as opposed to real imagery), and did so on simplistic scenes with limited variability.  As a result, it is unclear how well their methods would perform on more complex and realistic synthetic imagery, as well as real-world imagery.    


\subsection{Contributions of this work}
\label{subsec:introduction_contributions}

In this work we perform the first investigation of meta-simulation for real-world overhead imagery. We compare three meta-simulation methods: two existing methods, and one novel method.  Given the unique challenges of overhead imagery, we found that two existing methods were well-suited for the overhead imagery meta-simulation task (see Sec. \ref{subsec:related_work_MS}): direct regression (DR)\cite{Du2021Auto-Tuned}, and MS2\cite{Devaranjan2020}.  DR is a relatively simple approach that has recently been applied to meta-simulation \cite{Du2021Auto-Tuned}, and is suitable for overhead imagery. MS2 \cite{Devaranjan2020} recently achieved state-of-the-art performance, outperforming its predecessor Meta-Sim \cite{Kar2019a} on several meta-simulation tasks. 

\begin{figure}
    \centering
    \includegraphics[height=5.0cm]{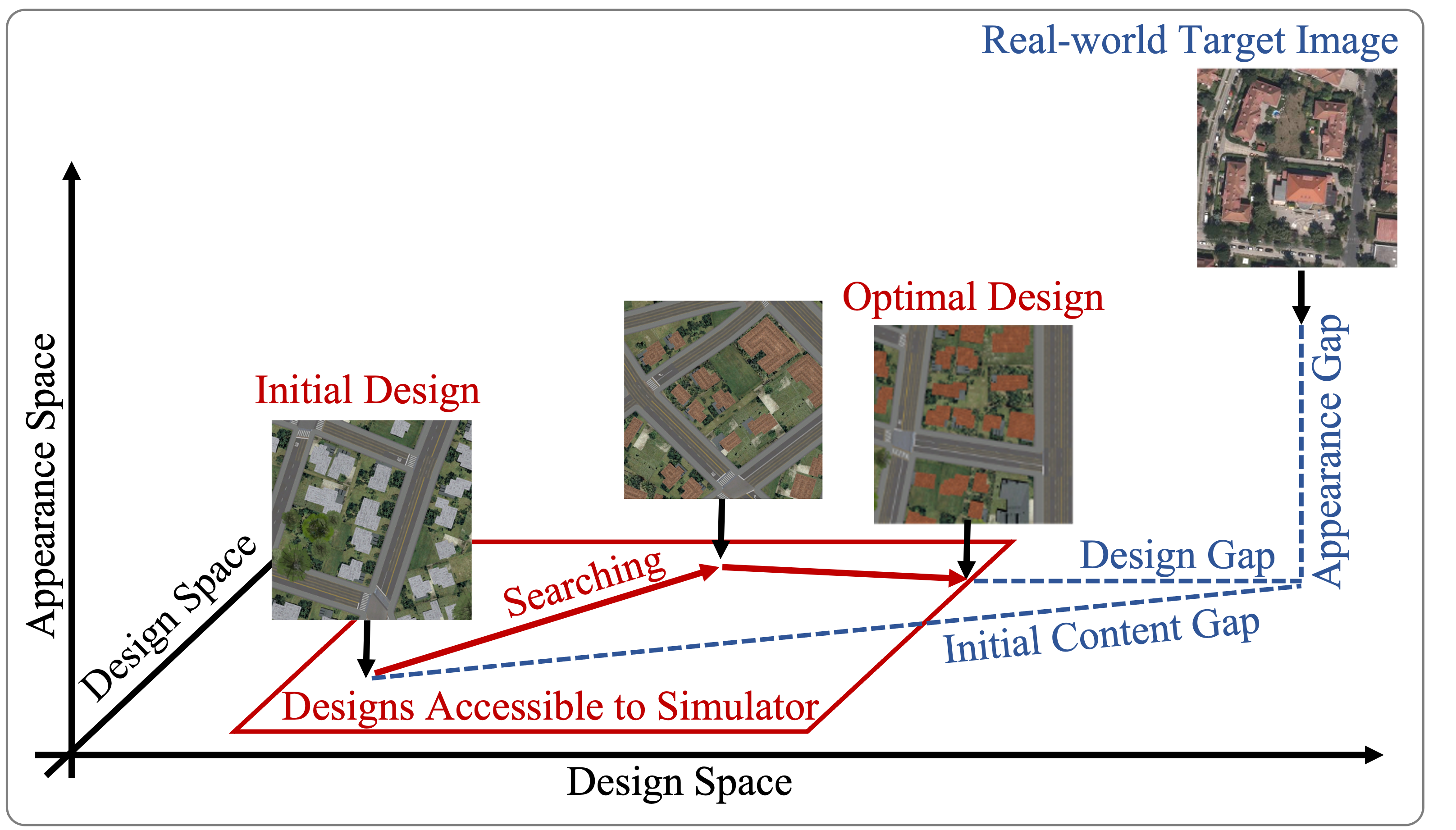}
    \caption{Sim-to-real gaps that impact the meta-simulation models are illustrated: (i) by searching among the designs (e.g., rooftop textures) accessible to the simulator, the content gap can be reduced, and the remaining content gap  after the optimal design is selected is called the design gap; (ii) the appearance gap exists due to the limitation of computer graphics, and it cannot be reduced by the meta-simulator.}
    \label{fig:illustration_of_appearance_and_design_gaps}
\end{figure}

We evaluated these two meta-simulation methods on several controlled experiments in which we isolate and examine the impact of two major real-world factors on meta-simulation models: (i) the presence of sim-to-real \textit{design} gaps, and (ii) the presence of sim-to-real \textit{appearance} gaps. 

As is presented in Fig. \ref{fig:illustration_of_appearance_and_design_gaps}, regarding (i) a design gap is the remaining content gap between the optimal synthetic scene design and the content in the real-world target imagery. As is mentioned in Sec. \ref{sec:introduction}, meta-simulation models aim to reduce the content gap by optimizing the design of virtual scene. However, in real applications, the content gap is unlikely to be completely eliminated. This may occur, for example, if the target imagery contains a rooftop with a texture that cannot be perfectly matched by any available to the meta-simulator.  Therefore, there is a "gap" between the designs available to the meta-simulator and those present in the target scene.  Given the variety of real-world scenes, design gaps are likely present whenever meta-simulating on any real-world target imagery.  In these cases, the meta-simulator will ideally identify the design that results in synthetic imagery that is visually most similar to the real-world imagery. 

The second factor is the presence of a sim-to-real appearance gap, which are typically low-level gaps that cannot be (directly) influenced or reduced by the meta-simulator's effectiveness, and as we show, tend to make meta-simulation much more challenging.   Significant sim-to-real gaps are present in current synthetic overhead imagery simulators, and therefore this is an important factor to consider when evaluating meta-simulation models.


We incrementally introduce the two aforementioned factors into our meta-simulation problems in order to isolate their impacts on the meta-simulation models. Our experiments reveal that existing meta-simulation approaches work relatively well when there are no sim-to-real gaps, however, both MS2 and DR degrade, to varying degrees, as these gaps are introduced.  Most importantly, both methods perform much more poorly when applied to real-world overhead imagery, which we hypothesize exhibits both significant appearance and design gaps.  MS2 - a state-of-the-art meta-simulation method - also suffers from several additional limitations.  First, it requires a large number of simulations each time meta-simulation is performed, which quickly becomes impracticable if meta-simulation must be repeated (e.g., for unique collections of target imagery). This cost is exacerbated if the simulator is relatively slow, as is the case in overhead imagery.  These limitations are described in greater detail in Sec. \ref{subsec:problem_setting_meta_simulation_2} and \ref{subsec:problem_setting_direct_regression}. 

To address the limitations of existing approaches, we propose Neural-Adjoint Meta-Simulation (NAMS), which is more effective for meta-simulation on real-world overhead imagery, and has better computational complexity.  NAMS relies upon training a DNN to model the simulator, yielding a differentiable function that relates synthetic imagery to its design.  Using this model, we can then use gradient descent to rapidly search for design parameters that yield synthetic imagery that match some given target imagery.  NAMS is therefore \textit{amortized}: it still requires an initial random sampling of simulations to perform meta-simulation, however, this simulation process only needs to be done once, after which NAMS can be used to rapidly perform design inference for new domains.  In our experiments we find that NAMS outperforms MS2 (and DR) when using the quantity of simulations required by MS2 \textit{for just a single meta-simulation}. The contribution of this work is briefly summarized below:  

\begin{enumerate}
    \item \textit{A performance comparison of modern meta-simulation techniques for overhead imagery tasks.} We compare the effectiveness of several meta-simulation models for complex and challenging synthetic and real-world overhead imagery. To our knowledge we are the first to evaluate meta-simulation methods for real-world or (complex) synthetic overhead imagery.      
    
    \item \textit{We investigate the behavior of meta-simulation methods under several important real-world scenarios.} We show that meta-simulation methods perform very differently depending upon whether (i) there is an appearance gap between the synthetic and real-world imagery, and (ii) whether the target imagery is within the design space of the simulator.
    
    \item \textit{Neural-Adjoint Meta-Simulation (NAMS), a novel meta-simulation model.} NAMS achieves similar performance to other existing methods, but it is better-suited to the challenges of meta-simulation in overhead imagery.  In particular, it performs better on real-world overhead imagery, and scales much better with the number of unique meta-simulation tasks that must be performed.

\end{enumerate}

The remainder of this paper is organized as follows: Section \ref{sec:related_work} discusses related work; Section \ref{sec:problem_setting} discusses the meta-simulation problem formulation and the existing methods; Section \ref{sec:nams_description} discusses the proposed NAMS methodology; Section \ref{sec:experimental_design} and \ref{sec:experimental_results} discuss our experiments and results; and Section \ref{sec:conclusion} presents our conclusions.

\section{Related work} 
\label{sec:related_work}

\subsection{Training models with synthetic data} \label{sec:related_work_syn}
The exploration of synthetic training data as an alternative to real-world dataset collection and annotation has grown rapidly in recent years. Synthetic training data has been studied for a variety of tasks, such as segmentation and planning in driving \cite{Richter2016,Wrenninge2018,Prakash2019,Gaidon2016,Dosovitskiy2017,Alhaija2018,Ros2016}, indoor scene segmentation \cite{Zhang2017,Armeni2019,Handa2016,McCormac2017,Savva2019,Wang2017,Wu2018}, robotic control \cite{Tassa2018,Todorov2012,Sadeghi2016,Brockman2016}, optical flow estimation \cite{Butler2012}, \cite{Shugrina2019}, home robotics \cite{Puig2018,Kolve2017,Gao2019}, surveillance system design and evaluation \cite{Qureshi2008}, \cite{Taylor2007}, \cite{Qureshi2007}, and more.

Synthetic \textit{overhead} imagery has also been recently explored for object detection \cite{Shermeyer2021,Hu2021,Xu,Han2017,Ward2018} and segmentation \cite{Kong2020}, respectively. In this work, we adopt the Synthinel-1 simulation approach \cite{Kong2020}, because it is the only method thus far that features an research-accessible simulator, and software to aid the generation of synthetic overhead imagery.  However, as is stated in \cite{Kong2020}, it takes approximately 1 minute to create 36 synthetic overhead images, covering 1 square km of the ground on a standard desktop, which is substantially slower than processes employed for synthetic natural imagery generation. Please refer to \cite{Kong2020} for more details about the simulator. In this work, the downstream task is building segmentation, and the main challenge is to choose building features that are appropriate for real-world target imagery. This is a challenging problem because of the substantial domain gap (noted in \cite{Kong2020}), and the substantial visual variability in buildings across different geographic regions.

\subsection{Meta Simulation Methods}
\label{subsec:related_work_MS}

Many meta-simulation methods are presented as solutions of the aforementioned meta-simulation task \cite{Ruiz2019LEARNING,Behl2020AutoSimulate:,Kar2019a,Devaranjan2020,Mansinghka2013Approximate,Kulkarni2015Picture:,Yildirim2015,Louppe2019Adversarial,Ganin2018Synthesizing,Mellor2019Unsupervised}, however, most of them cannot be directly applied to our meta-simulation task on overhead imagery, due to three major unique properties of overhead imagery with respect to natural imagery: the presence of discrete and non-ordinal design parameters (e.g., selection of textures from a bank); the absence of labels in the target domain; and the large sim-2-real visual gap between synthetic and real overhead imagery.  We next summarize existing methods, and how each are rendered ineffective or inapplicable due to these factors. 

In \cite{Kulkarni2015Deep}, a neural network with an encoder-decoder structure is used to learn the design parameters from target images, however, the networks are designed for continuous parameters and do not support discrete/categorical parameters natively, which may be impractical in real-world overhead imagery applications. In \cite{Du2021Auto-Tuned}, a searching method is implemented, to iteratively decide whether the given parameters are higher or lower than the true parameters in the target data, and shift the parameters to approach the target parameters. However, for categorical parameters, there is no intrinsic ordering to different categories, and such a method also does not work for categorical parameters. 

In \cite{Ruiz2019LEARNING}, \cite{Behl2020AutoSimulate:}, Reinforcement Learning (RL) strategies are adopted for the searching of the target design parameters, with a bi-level optimization problem formulation. The lower level optimizes the variables in the downstream model, given the synthetic data generated with certain design parameter values, while the upper level optimizes the design parameter values, by maximizing the performance of downstream algorithms over a validation target dataset. RL techniques are applied to solve the aforementioned non-differentiable optimization. However, these methods rely on a set of labeled target data to form a validation dataset, which is very likely not available in real-world overhead imagery applications.

Other RL-based methods do not rely on labeled target data \cite{Louppe2019Adversarial,Ganin2018Synthesizing,Mellor2019Unsupervised}, where adversarial learning strategies are applied. Discriminators are trained to measure the difference between synthetic data and the target data. The design parameter values are then optimized by maximizing the capability to fool the discriminator, to avoid maximizing a validation loss. Therefore, these methods rely on only unlabeled target data for training the discriminator. However, as is discussed in \cite{Kar2019a}, learning with discriminators is known to suffer from mode collapses. Besides, due to the large sim-to-real gap in the overhead imagery applications, discriminators can easily distinguish synthetic image and real world target images, making the learning of the parameters fail due to vanishing gradients.

There are some existing methods that are potentially applicable to overhead imagery.  One such method is direct regression (DR) \cite{Du2021Auto-Tuned}, where a CNN model is used to directly infer design parameters from images. The model is trained using a set of synthetic images, paired with their corresponding design parameters. We include this method in our experiments because it is applicable, and to demonstrate that such a simple and straightforward approach does not work well on real-world overhead imagery.  Another class of applicable approaches evaluate the difference between synthetic data and the target data by using hand-crafted distance metrics between synthetic and real-world imagery (as opposed to learned adversarial distances) and hence \cite{Kar2019a,Devaranjan2020,Mansinghka2013Approximate,Kulkarni2015Picture:,Yildirim2015}. For example, Maximum Mean Discrepancy is used in \cite{Kar2019a} and feature distribution matching is used in \cite{Devaranjan2020}.  Among these methods, MS2\cite{Devaranjan2020} recently achieved state-of-the-art performance and therefore we include it in our experiments.  We use our experiments to demonstrate that both DR and MS2 suffer from unique limitations that undermine their performance on real-world overhead imagery. Additionally, MS2 exhibits computational properties that make it scale poorly to multiple target domains, which is discussed further in Sec. \ref{subsec:problem_setting_meta_simulation_2}. 

\subsection{The Neural Adjoint}
\label{subsec:related_work_NA}

The Neural-Adjoint (NA) is a recently-proposed method to solve ill-posed inverse problems \cite{Ren2020}, which is an extension of other similar methods (e.g., \cite{peurifoy2018nanophotonic,Gomez-Bombarelli2018}). Our meta-simulation task can be framed as an ill-posed inverse problem, where we observe some data, and then we must identify some hidden parameters (e.g., design parameters) that will give rise to the observed data. In \cite{Ren2020} the NA achieved superior overall performance compared to other contemporary approaches on a benchmark of nonlinear inverse problems, and therefore we adopt it here.

\section{Problem Setting and Existing Methods} 
\label{sec:problem_setting}
In this section we define the meta-simulation problem, and introduce some existing methods utilized in our experiments.  In our context, the goal of meta-simulation is to train some type of supervised recognition model (e.g., segmentation, object detection) for overhead imagery, denoted $y = f_{\gamma}(x)$, where $x$ is some input image, $y$ are the labels corresponding to the imagery (e.g.,pixel-wise labels, or object bounding boxes), and $\gamma$ represents the model parameters (e.g., weights of a DNN).  We assume the availability of some set of labeled \textit{source domain} data $(X^{S},Y^{S}) = \{ (x^{S}_{i},y^{S}_{i})\} _{i=1}^{N^{S}}$ that can be used to infer $\gamma$ (i.e., train the task model). It is assumed that $x^{S}_{i} \sim p^{S}$, where $p^{S}$ is the distribution of the source domain imagery.  We then wish to apply our trained model, $f_{\gamma}$, to a collection of unlabeled \textit{target domain} imagery $X^{T} = \{x^{T}_{i}\}_{i=1}^{N^{T}}$ where $x^{T}_{i} \sim p^{T}$, where $p^{T}$ is the target domain distribution.  We assume in general that $p^{S} \neq p^{T}$ so that the trained model is likely to perform poorly when applied to predict the labels for any $x^{T}_{i} \in X^{T}$.  

This scenario arises frequently in remote sensing applications where we wish to apply a trained model to new imagery that was collected under different conditions than the training imagery: e.g., a novel geographic locations, weather conditions, times-of-day, and imaging hardware.  Recent research has indicated that such differences across image collections often cause significant reductions in the performance of DNNs, even when using a large and diverse training dataset \cite{Kong2020,Maggiori2017}. 

One strategy to overcome this problem is to generate a set of synthetic imagery that resembles the target domain imagery, and then use it to help train $f_{\gamma}$ so that it performs better when applied to $X^{T}$.  More formally, assume that we have some set of synthetic data $(X^G,Y^G)=\{(  x^{G}_{i}, y^{G}_{i}) \}_{i=1}^{N^{G}}$ that has been generated by some process, denoted $G(d,\zeta)$, where $(x^{G}_i,y^{G}_i) = G(d_{i},\zeta_{i})$ and $d_{i} \sim p^{d}$ and $\zeta_{i} \sim p^{\zeta}$.  Here $G$ refers to a simulation engine for generating synthetic imagery, and $d \in \mathcal{D}$ encodes the contents of the synthetic imagery that is provided to the simulator: e.g., the objects in the scene, their locations, and their appearance.  We call $\mathcal{D}$ the design space, encompassing all scene content \textit{that we can control}.  Note that $d$ may be composed of continuous or discrete values, or a mixture.  The distribution $p^{d}$ therefore controls the properties of the synthetic data, and must be set by the designer. The vector $\zeta$ models the properties of the synthetic imagery that vary randomly each time synthetic imagery is generated, but that are not controlled by $d$. For simplicity we will often omit $\zeta$ when discussing $G$, since it is usually immaterial.   

Recent work has demonstrated that synthetic overhead imagery can be highly beneficial for training recognition models for overhead imagery \cite{Kong2020,Ward2018,Shermeyer2021,Xu,Hu2021,Han2017}, e.g., when training on an augmented dataset $(X^{S},Y^{S}) \cup (X^{G},Y^{G})$ \cite{Kong2020,Shermeyer2021,Hu2021,Han2017}.  However, choosing  $p^{d}$ is costly and time-consuming task that greatly undermines the value of the synthetic data.  \textit{The goal of meta simulation is to automatically infer a setting of $p^{d}$ that maximizes the effectiveness of the resulting set of synthetic imagery, only using the unlabeled target domain data, $X^{T}$}.  

\subsection{Challenges and Modeling Strategies} 
\label{subsec:challenges_of_meta_simulation}

Existing meta-simulation models generally work by trying to align the distribution of the synthetic imagery, denoted $p^{G}$, with the distribution of the target domain imagery $p^{T}$.  Mathematically, they generally try to solve the following optimization problem
\begin{equation} \label{eq:meta_simulation_alignment}
    p^{d*} = \argmin_{p^d} \mathcal{L}(p^{G},p^{T})
\end{equation}
where $p^{d*}$ is the distribution of $d$ that minimizes the difference, measured by $\mathcal{L}$ between the synthetic and target data distributions.  Note that $x^{G} = G(d)$ where $d \sim p^{d}$, and therefore $p^{G}$ depends directly upon $p^{d}$, the distribution of $d$, which is controlled by the designer.  

This is a challenging optimization problem problem due to several factors.  First, it is difficult to model $p^{G}$ and $p^{T}$ due to the high dimensionality of the imagery.  Then alignment of the distributions is difficult because $p^{G}$ depends upon $d$ through the function $G$, which is a complex and non-differentiable function, which may also be computationally costly to evaluate. Finally, some properties of the synthetic imagery vary randomly due to $\zeta$, making it unlikely to achieve a perfect alignment of distributions, and adding noise to any estimates of the distributions or their alignment.    

To simplify the meta-simulation problem, recent methods extract lower-dimensional features of the imagery using a pre-trained DNN \cite{Kar2019a,Devaranjan2020}, such as the ResNet model \cite{he2016deep}.  In this case $x^{T}$, $x^{S}$, and $x^{G}$ above would represent feature vectors of imagery, rather than the original imagery.  These substitutions can be made without any loss of generality in the problem description, or descriptions of the methods.  




\subsection{Meta-Sim2 (MS2)}
\label{subsec:problem_setting_meta_simulation_2}

MS2 \cite{Devaranjan2020} was proposed as an improvement to the Meta-Sim method \cite{Kar2019a}.  MS2 models $p^{d}$ with different statistical assumptions depending on the type of variable $d$, for example, the authors use a multinomial distribution for categorical values of $d$, which has the advantage that it has continuous distribution parameters, denoted $\theta$. They then approximate $p^{G}$ and $p^{T}$ using kernel density estimation, with the approximations denoted by $\hat{p}^{G}$ and $\hat{p}^{T}$, respectively. Then the authors formulate the optimization in Eq. (\ref{eq:meta_simulation_alignment}) as
\begin{equation}
    \label{eq:MS2}
    \min_{\theta} \textit{KL}[\hat{p}^{T}||\hat{p}^{G}],
\end{equation}
\begin{equation}
    \label{eq:MS2_2}
    \min_{\theta} \mathbb{E}_{x^G \sim p^{G}} [\log \hat{p}^{G}(x^G) - \log \hat{p}^{T}(x^G)],
\end{equation}
where $\mathbb{E}$ represents the expectation operator, and $\textit{KL}$ represents the Kullback-Leibler divergence between the two distributions.  Using the relationship that $x^G=G(d)$ and $d \sim p^{d}$ we can rewrite the loss as 
\begin{equation}
    \label{eq:MS2_3}
    \min_{\theta} \mathbb{E}_{d \sim p^{d}} [\log \hat{p}^{G}( G(d) ) - \log \hat{p}^{T}( G(d) )],
\end{equation}
This objective is still difficult to optimize because of the (non-differentiable) simulation engine, $G$, present in its computation. To optimize this objective the REINFORCE estimator is utilized, whereby the gradient of the loss in Eq. (\ref{eq:MS2_3}) is approximated with
\begin{multline}
    \label{eq:MS2_4}
    \nabla_{\theta}\mathcal{L} \approx \\ 
    \frac{1}{N} \sum_{i=1}^N (\log \hat{p}^{G}( G(d_{i}) ) - \log \hat{p}^{T}( G(d_i)) 
    \nabla_{\theta} \log p^{d}(d_{i}).
\end{multline}

In  \cite{Devaranjan2020} the authors use $N=500$ to obtain an accurate estimate of the gradient, which must be simulated with $G$ with $d \sim \hat{p}^{d}$ using the current estimate of $\theta$.  Once the gradient is obtained, it is used to adjust $\theta$, and the kernel density estimation is used to re-estimate $p^{G}$.   This process is repeated until convergence of the loss in Eq. (\ref{eq:MS2_3}).  In \cite{Devaranjan2020} the authors used 200 iterations.  A major drawback of MS2 is its computation time, due to the number of simulations that must be run per iteration of gradient descent.  Furthermore, this process must be repeated each time meta-simulation must be performed (e.g., if a new domain is encountered).  For this reason MS2 scales poorly with the number of meta-simulation runs. These problems are further exacerbated if the simulator is relatively slow, such as in the case of overhead imagery.  Collectively these problems make MS2 computationally intensive for overhead imagery applications.  In Section \ref{sec:experiment_computation_time} we provide a comparison of computation time between different meta-simulation methods considered in this work. 

One other important limitation of this approach is its dependence upon kernel density estimates for $\hat{p}^{T}$, which requires a large number of samples of real imagery from the target domain (e.g., $N=500$ in \cite{Devaranjan2020}), which may not always be available due to the cost of overhead imagery.  

\subsection{Direct Regression (DR)}
\label{subsec:problem_setting_direct_regression}

Another way to solve Eq. (\ref{eq:meta_simulation_alignment}) is to infer the appropriate design for each $x^{T}_{i} \in X^T$ independently, resulting in a collection of designs, given by $\hat{D}^T=\{\hat{d_{i}}(x_{i}^{T})\}$, where each $\hat{d}_{i} \in \hat{D}^T$ corresponds to one target domain instance. The designs in $D^T$ can then be used to estimate $p^{d}$, or be sampled directly to generate synthetic training imagery. This is the approach taken by direct regression (DR), which has been previously applied in \cite{Du2021Auto-Tuned}.    

In direct regression we train a model of the form $\hat{d} = f_{\lambda}^{\textit{DR}}(x)$ that directly predicts the design of a given image.  This model is trained using a collection of $N^{\textit{DR}}$ randomly-sampled synthetic training imagery, $D^{\textit{DR}} = \{(G(d_i),d_i) \}_{i=1}^{N^{\textit{DR}}}$ where $d_{i} \sim U$.  Here $U$ is a uniform distribution over $\mathcal{D}$, the domain of $d$, so that $D^{\textit{DR}}$ is representative of all possible designs, and therefore the trained regression model will be accurate for all possible designs within the design space. The regression model is then trained to satisfy the following objective
\begin{equation} \label{eq:direct_regression_optimization}
    \min_{\lambda} \mathbb{E}_{d \sim U} [\mathcal{L}( f_{\lambda}^{\textit{DR}}(G(d)), d)], 
\end{equation}
where $\mathcal{L}$ is some loss function. In this work we use mean-squared error for the loss, and we estimate the expectation in Eq. (\ref{eq:direct_regression_optimization}) using $D^{\textit{DR}}$.  As discussed in \ref{subsec:challenges_of_meta_simulation}, $x$ in this case represents features of imagery extracted from a pre-trained DNN, rather than raw imagery, to reduce the dimensionality of the problem.  We use a DNN for $f^{\textit{DR}}_{\lambda}$.    

This approach has the advantage that it relies upon training a standard regression model, making the training process relatively simple and fast.  Furthermore, it  only requires one initial collection of synthetic imagery for training, rather than requiring large quantities of simulation each time a new domain is encountered like MS2. However, this approach has the limitation that it can make highly inaccurate predictions if the real-world target imagery is visually novel compared to the synthetic training imagery \cite{Du2021Auto-Tuned}.  This can occur, for example, if there are (low-level) visual domain gaps between the synthetic and real-world imagery, or if the real-world imagery contains content that cannot be instantiated with the the simulator, $G$.


\section{Neural-Adjoint Meta-Simulation (NAMS)} 
\label{sec:nams_description}

Neural-Adjoint Meta-Simulation (NAMS) combines the advantages of DR and MS2 into a single model. Similar to DR, NAMS only needs to be trained once, after which it can be applied to rapidly infer designs for new target domains.  NAMS also only requires a few target domain images to infer a design (e.g., we use 9 images in our experiments).  However, NAMS is more robust than DR to visual gaps between the synthetic and real-world imagery, leading to much better design inference on real-world problems. As we show in Sec. \ref{sec:experimental_results}, DR can fail unpredictably and dramatically, while MS2 and NAMS can perform more reliably.  

\begin{figure}
    \centering
    \includegraphics[height=8.5cm]{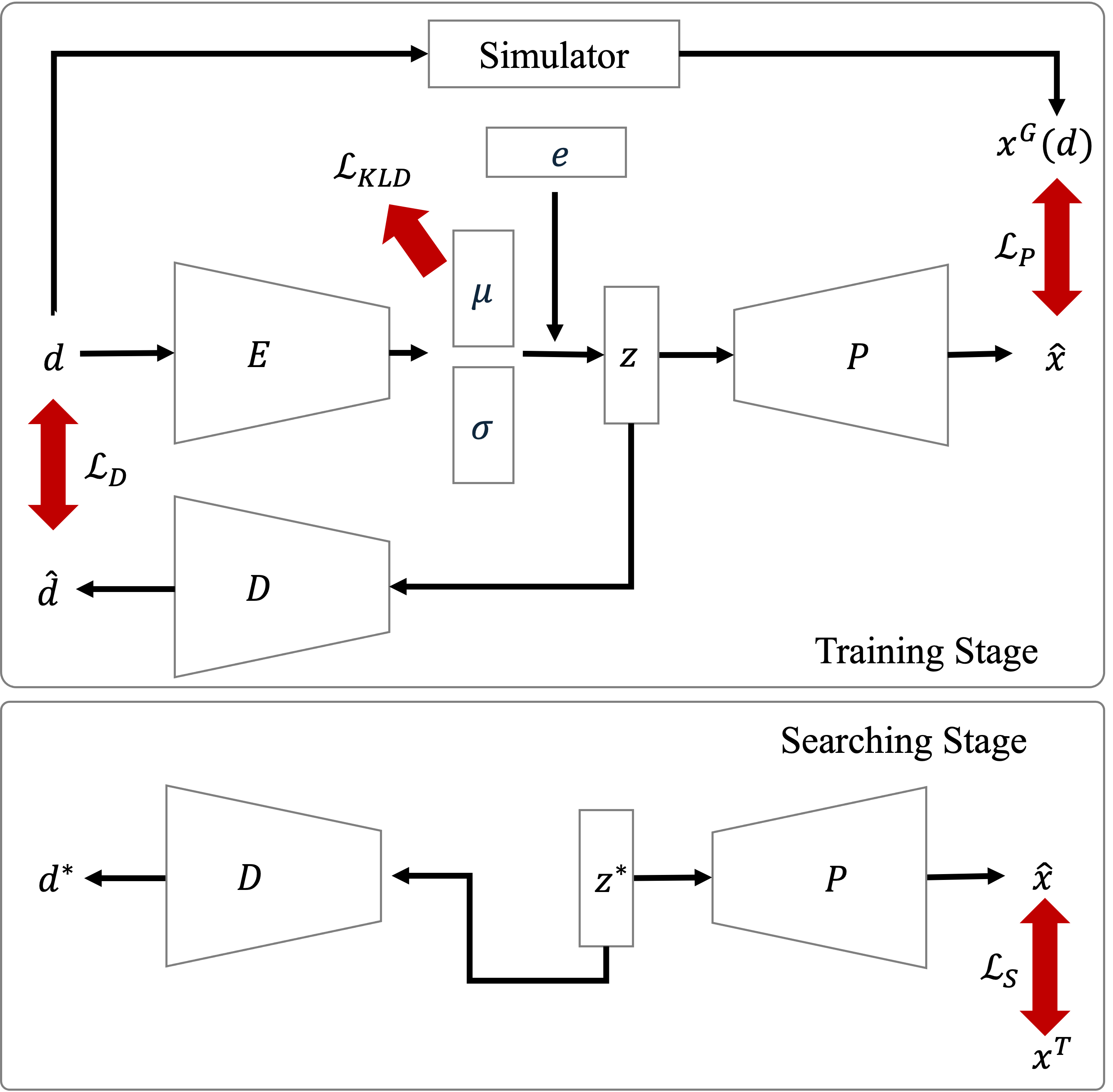}
    \caption{Overview of NAMS in two stages: in the training stage, mappings between continuous embedding $z$ and the design $d$ using network $E$ and $D$ are trained, together with an adjoint neural network $P$, by using synthetic data $d$ and $x^G(d)$; in the searching stage, the target continuous embedding $z^*$ is searched based on the target data $x^{T}$ and the corresponding design $d^*$ is reconstructed.}
    \label{fig:NAMS}
\end{figure}

\textbf{Model Overview.} A diagram of NAMS is presented in Fig. \ref{fig:NAMS} (top), where the functions $E$, $P$, and $D$ are all DNN with parameters given by $\theta_{E}$, $\theta_{P}$, and $\theta_{D}$ respectively.  A major obstacle of meta-simulation is the black-box graphics simulator, which makes it difficult to optimize Eq. (\ref{eq:meta_simulation_alignment}). NAMS addresses this issue by using two DNNs to model the simulator, given by $\hat{x} = P(E(d))$ in Fig. \ref{fig:NAMS}.  With a DNN-based simulator we can perform gradient descent with respect to $d$ to minimize the visual differences between our simulated imagery and a specific target domain image.  Similar to DR, we can identify the design for each available target domain image, resulting in a collection of estimated designs that can then be used to construct $p^{d}$ to minimize the meta-simulation objective in Eq. (\ref{eq:meta_simulation_alignment}). 

A second challenge of meta-simulation is the presence of discrete or non-ordinal entries of $d$, which can undermine gradient descent.  To mitigate this problem, we learn a \textit{continuous} embedding of the designs, denoted $z$, such that similar embeddings have similar-looking synthetic imagery. We use a stochastic embedding, where $z$ is sampled from a normal distribution with parameters that depend upon the $d$.  Mathematically, we have $z \sim \mathcal{N}(\mu,\sigma)$ where $(\mu,\sigma) = E(d)$.  In the NAMS model, we use $z = \mu + e \odot \sigma$, where $e \sim p^{e}$ is a normal random variable with diagonal covariance and the operator $\odot$ performs element-wise multiplication of two vectors, following \cite{kingma2013auto}. The randomness of $z$ is intended to model visual features of the synthetic and real imagery that are not controlled by $d$, and therefore vary across imagery even while $d$ is fixed (e.g., the $\zeta$ parameter in Sec. \ref{sec:problem_setting}).  

When inferring the setting of $d$ that corresponds to some target image, we first perform gradient descent with respect to $z$. We then decode the optimized value, $z^{*}$ using $D$ to get $d^{*}$ as illustrated in Fig. \ref{fig:NAMS} (bottom).  We next present the technical details of NAMS.


\textbf{Model Training.}  During training we minimize the following overall loss given by 
\begin{equation} \label{eq:nams_overall_loss}
    \mathcal{L}_{\textit{NAMS}} = \lambda_{P} \mathcal{L}_{P} + \lambda_{D} \mathcal{L}_{D} + \lambda_{\textit{KLD}} \mathcal{L}_{\textit{KLD}},
\end{equation}
where $\lambda_{P}$, $\lambda_{D}$, and $\lambda_{\textit{KLD}}$ represent the weight of each loss term during training.  The first loss, $\mathcal{L}_{P}$, encourages the model $\hat{x}=P(E(d)) \sim x^G(d)$ to make accurate predictions of image features, given some design for the imagery.  Mathematically, we have
\begin{equation}  \label{eq:nams_predictor_loss}
    \mathcal{L}_{P}(\theta_{E},\theta_{P}) = \mathbb{E}_{d \sim U, \zeta \sim p^{\zeta}, e \sim p^{e}} [ \mathcal{L}(P(E(d)),G(d,\zeta))],
\end{equation}
where and $\zeta$ is the random noise vector used as input to the true simulator, $G$. Here $\mathcal{L}$ represents some measure of error, and we use mean-squared error. Since both $E$ and $P$ are DNNs, we can infer their parameters using gradient descent. 

The second loss, $\mathcal{L}_{D}$, encourages the model to obtain accurate decoding of the latent vector $z$ into its corresponding design.  Mathematically we have
\begin{equation}  \label{eq:nams_decoder_loss}
    \mathcal{L}_{D}(\theta_{E},\theta_{D}) =  \mathbb{E}_{d \sim U, e \sim p^{e}} [ \mathcal{L}(D(E(d)),d)].
\end{equation}
Again, since $E$ and $D$ are both DNNs, this loss can be minimized via gradient descent.  

The third loss, $\mathcal{L}_{\textit{KLD}}$, encourages the design embeddings to be smooth and independent. Mathematically we have
\begin{equation}  \label{eq:nams_kl_loss}
    \mathcal{L}_{\textit{KLD}}(\theta_{E}) = \mathbb{E}_{d \sim U, e \sim p^{e}} [ \textit{KL}(p^{z},p^{e})],
\end{equation}
where $p^{e}$ is a multivariate normal distribution with zero mean and diagonal covariance.  In all three loss equations above, we replace the expectation operators with sample estimators when optimizing on real-world data.   In section \ref{sec:experimental_design} we will discuss the specific implementation details of the DNNs in NAMS, and our hyperparameter settings. 

\textbf{Inference of Designs for Data Generation.}  In the NAMS searching stage, we use the predictor $P$ to solve the optimization in Eq. (\ref{eq:meta_simulation_alignment}) in $z$ space, similar to the DR method, by inferring the appropriate design for each $x^{T}$ independently, as shown in Fig. \ref{fig:NAMS}. With the trained predictor, $P$, we can directly infer synthetic image features $P(z)=\hat{x} \sim x^G(d)$, given the embedding $z$. We can therefore substitute this into Eq. (\ref{eq:meta_simulation_alignment}), so that the searching optimization can be rewritten as:
\begin{equation}
    \label{eq:na}
    \min_{z}\mathcal{L}_{S}[P(z), x^T].
\end{equation}
Note that, $P$ is a differentiable neural network. As a result the optimization can be efficiently solved using a gradient-decent-based searching method. For this, we used the recently-proposed Neural-Adjoint approach \cite{Ren2020}, which freezes the network weights of $P$ and treats the embedding $z$ as the parameters and then performs back-propagation with respect to the loss $\mathcal{L}_S$ to solve the optimization problem in Eq. (\ref{eq:na}).

\textbf{Initialization and Majority Voting.}  Due to the non-linearity of the problem, the search space is likely to contain many local minima. Hence, for each target image, we propose to search from $M$ random initial points uniformly distributed within a range of $\pm k \sigma_j$ in $j^{th}$ dimension in the space of $z$ and perform a majority voting among $S$ searching results with the smallest losses, which are considered near the global minimum. Here, the hyper-parameters $M$ is set to 1000, $k$ is set to 3 and $S$ is set to 50 for all of our experiments, and $\sigma_j$ is estimated as the standard division of the training data in $j^{th}$ dimension of $z$. Together the optimization that we solve during the search phase becomes:
\begin{equation}
    \label{eq:z}
    \begin{split}
        &z^* = V_S[\{z^{(i)*}\}_{i=1}^M], \\
        &\text{s.t.} \quad z^{(i)*} = \argmin_{z^{(i)}} L_2[P(z^{(i)}) - x^{T}], \\
    \end{split}
\end{equation}
where the element-wised $L_2$-norm $L_2$ is used to measure the distance in feature space, and $V_S$ stands for the majority voting operation among $S$ best results.

Once we infer $z^*$, we can then use the decoder to obtain the corresponding design, $d^*=D(z^*)$.
This searching process is repeated for each individual real-world target image, resulting in a dataset of optimized $d^*$ values, which we can use as a population to simulate data in the simulator. Similarly, we can treat the population of designs as a sampling distribution for designs of our synthetic training data for a downstream task.

\textbf{Augmentation and Feature Averaging.} In addition, to reduce the impact of the randomness caused by $\zeta$, data augmentation and averaging can be applied. In all of our experiments, both in the training and searching stages, we take 9 images with the same design $d$ to form a mini-batch each time. For each image, we create 8 augmented images with four rotations $\{0^{\circ},90^{\circ},180^{\circ},270^{\circ}\}$ and a horizontal flip. Then, we average the features of these 72 images in the mini-batch. Hence, $x^{T}$, $x^{S}$, and $x^{G}$ above represent the average features of 72 images with the same design $d$.

\textbf{Diversify with Rejection.} In real applications, we randomly discard some proportion of the optimized designs, and replace them with a uniformly selected design (the default non-meta-simulation design approach). This has the effect of diversifying the simulated training data for the downstream task, and mitigates risks of over-fitting to (possibly erroneous) $z^*$ estimates. To achieve this, we use a weight $r$ in $(0,1)$ to control the probability of rejecting the selected design and replacing with a uniformly selected design, so that the probability to obtain rare designs will increase. The sampling method can be described as follows:
\begin{equation}
    \label{eq:d}
    d^* = 
    \begin{cases}
        D(z^*) ,           & u\geq r\\
        d \sim U,             &  u < r
    \end{cases},
\end{equation}
where $u$ is a uniformly sampled value in $(0,1)$, and note that $U$ stands for a uniform distribution over $\mathcal{D}$. 

We summarize the overall algorithm of the proposed NAMS method in Algorithm 1.
\begin{algorithm}[h!]
  \caption{ }
  \label{alg:1}
    \begin{algorithmic}
      \IF{Training stage}
        \STATE Input a training dataset $\{(d_i,x^G_{i}(d_i) ) \}_{i=1}^{N^G}$.
        \STATE Do augmentation, extract features and do averaging.
        \STATE Set weights $\lambda_D$, $\lambda_{KLD}$, and $\lambda_P$.
        \STATE Train \textit{E}, \textit{D}, \textit{P} with the optimization in Eq. (\ref{eq:nams_overall_loss}).
      \ENDIF 
      \IF{Searching Stage}
        \STATE Input a real-world target dataset $X^{T}$.
        \STATE Set hyper parameter $r$.
        \STATE Do augmentation, extract features and do averaging.
        \FOR{Each $x^{T} \in X^{T}$}
            \STATE Search for the design embedding $z^*$ by Eq. (\ref{eq:z}).
            \STATE Obtain the optimal design $d^*$ by Eq. (\ref{eq:d}).
        \ENDFOR
        \STATE Generate synthetic dataset $X^{G*}$ using all $d^*$ values in the simulator.
      \ENDIF
    \STATE Train the downstream task algorithm using $X^{G*}$.
    \end{algorithmic}
\end{algorithm}

\section{Experimental Datasets} 
\label{sec:experimental_datasets}
We evaluate three meta-simulation methods of satellite imagery for the task of building segmentation. Building segmentation is a challenging and widely-studied problem in remote sensing \cite{Huang2018,Maggiori2017,Demir2018}, making it a useful and representative problem on which to examine meta-simulation for overhead imagery. To support these experiments, we construct both real-world and synthetic overhead imagery datasets, as described next.

\subsection{Real-world Overhead Imagery.}  

We obtain real-world overhead imagery labeled with building footprints from DeepGlobe (DG) \cite{Demir2018} and Inria \cite{Maggiori2017}). These two popular benchmark datasets for building footprint segmentation feature $0.3m/pixel$ resolution RGB imagery collected over nine diverse cities across the world, encompassing $713 km^2$ of surface area.  These two datasets collectively provide a challenging and contemporary set of real-world overhead imagery for testing both our task models and our meta-simulation models.   In our experiments, all of the images are cropped into the same size $x\in R^{384\times384}$. 

\subsection{Synthetic Overhead Imagery.} 

We create synthetic overhead imagery $x\in R^{384\times384}$ using the only one research-accessible simulator from \cite{Kong2020}. First, we procedurally generate virtual cities with numerous features (e.g., road network topology, building types, etc). The content of each virtual city depends upon a large number of parameters. Some parameters are set to fixed values (e.g., lighting conditions), while others are sampled from distributions (e.g., object types and density). Please see the detailed parameter settings in \cite{Kong2020}. Then, synthetic satellite imagery is captured using a virtual camera (with $0.3m/pixel$ resolution) overlooking the virtual city. The simulation requires nearly 1.6 seconds per synthetic image on our hardware, which is similar to the speed stated in \cite{Kong2020}. 

\section{Experimental Design} 
\label{sec:experimental_design}


In our experiments, we assume that we have access to a set of real-world overhead imagery labeled with building footprints, for training building segmentation models. We use the DG dataset as this labeled source domain data $(X^{S},Y^{S})$. Our goal is to apply the models to different novel unlabeled target datasets, which may be visually dissimilar to the labeled source imagery (i.e., the same DG imagery in our case) used to train our task models. 

Due to the large difference between the training and target datasets, we use synthetic imagery to augment the training of the downstream building segmentation models. The ratio of the real-world training images and the synthetic augmenting images is set to 6:1, following \cite{Kong2020} for the best performance. 

To improve the efficacy of synthetic images, our meta-simulation problem aims to infer simulator design parameters that result in synthetic imagery that is most visually similar to the unlabeled target imagery. In our experiments, we infer two design parameters, $d_{flat} \in [0,1]^{40}$ and $d_{slope} \in [0,1]^{44}$, which controls “rooftop texture” of (i) flat-roofed buildings and (ii) sloped-roof buildings, respectively. For simplicity, we infer the dominant rooftop textures in each target image only. In other words, only one $d_{flat}$ value and one $d_{slope}$ value are inferred per target image. For consistency, in the simulator, we use the same $d_{flat}$ and the same $d_{slope}$ for all of the sloped-roof and flat-roofed buildings, respectively, in each virtual city. 

\begin{figure}
    \centering
    \includegraphics[height=5.6cm]{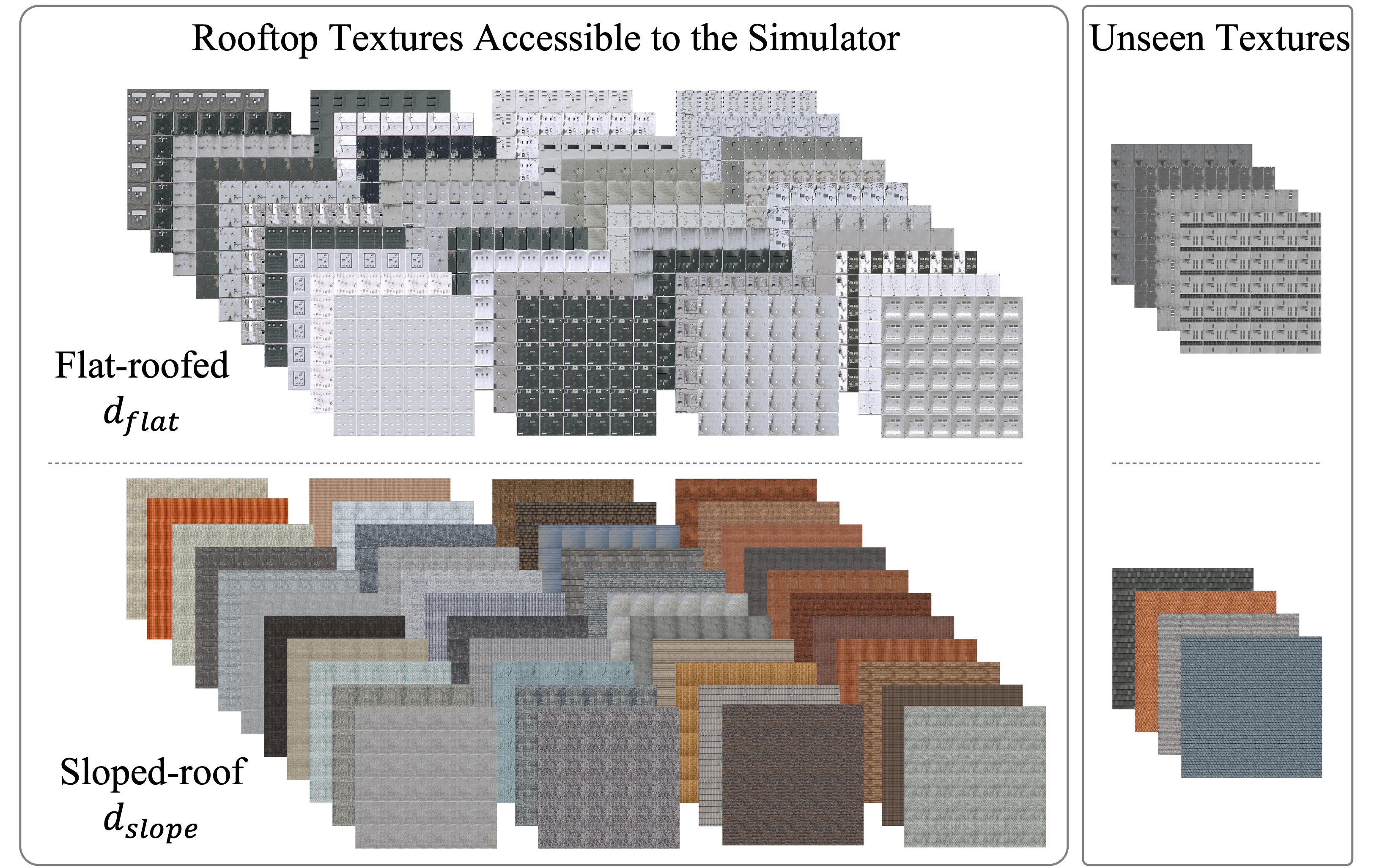}
    \caption{Rooftop textures of flat-roofed and sloped-roof buildings used in the experiments are shown.}
    \label{fig:exmaples_textures}
\end{figure}

The rooftop textures accessible to the model are shown in the left column of Fig. \ref{fig:exmaples_textures}.  We choose these two parameters because we hypothesize that building color and texture have a strong influence on the the building segmentation task. We use a one-hot encoding for these two parameters, making them challenging categorical and non-ordinal variables to infer i.e., the roof texture at index of one may be much more similar to the texture at index of forty than to the texture at index of two. Because we use building rooftop textures as the design parameter of interest, we only include target imagery in our test set with at least 10000 building pixels ($6.78\%$ of total) in it.  

\subsection{Experimental Scenarios}

\begin{figure}
    \centering
    \includegraphics[height=5.6cm]{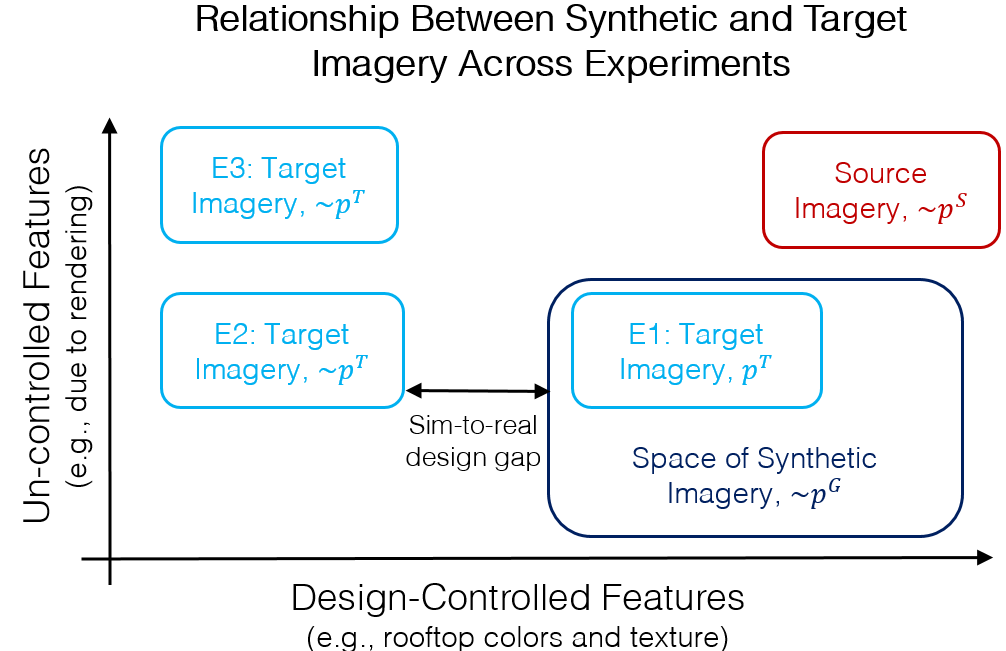}
    \caption{Illustration of the relationship between the different sets of imagery employed in our meta-simulation experiments.  For example, E1 denotes the set of target imagery employed for Experiment 1.}
    \label{fig:experimental_design_illustration}
\end{figure}

As is mentioned in Sec. \ref{subsec:introduction_contributions}, we examine the impact of sim-to-real design/appearance gaps to meta-simulation methods with three increasingly-challenging target scenarios, as shown in Fig. \ref{fig:experimental_design_illustration} and summarized in Table \ref{table:experimental_design_outline}. 

\setlength{\tabcolsep}{4pt}
    \begin{table}
        \begin{center}
        \caption{Experimental Design Outline}
        \label{table:experimental_design_outline}
            \begin{tabular}{ccccc}
                \hline
                \begin{tabular}[c]{@{}c@{}}Experiment \\ ID\end{tabular}  & \begin{tabular}[c]{@{}c@{}}Training Domain \\ Data, $X^{T}$ \end{tabular} & 
                \begin{tabular}[c]{@{}c@{}}Target Domain \\ Data, $(X^{S},Y^{S})$ \end{tabular}  & \begin{tabular}[c]{@{}c@{}} Appearance \\ Gap? \end{tabular} & 
                \begin{tabular}[c]{@{}c@{}} Design \\ Gap? \end{tabular} \\
                \hline
                E1  & Synthetic & Synthetic & No  & No \\
                E2  & Synthetic & Synthetic & No  & Yes \\
                E3  & Real      & Real      & Yes & Yes \\
                \hline
            \end{tabular}
        \end{center}
    \end{table}
    
\textbf{E1: Synthetic Target Data with No Design Gap} 

In the first scenario, we use imagery generated from our simulator with no design gap as target imagery. In other words, the target scene designs are accessible to the meta-simulator, and hence, the meta-simulator is able to perfectly match the target scene designs. This scenario is the easiest, and although it is unrealistic, it serves as a useful baseline to analyze the impact of adding in more realistic conditions.  To increase the reliability of the experiment, we do repeated trials using 4 target datasets, each containing 900 images collected from a synthetic city with a pair of randomly sampled $d_{flat}$ and $d_{slope}$ accessible to the meta-simulator.

\textbf{E2: Synthetic Target Data with a Design Gap }

In the second scenario, we again use synthetic target imagery, however, it is constructed using designs that are not accessible to the meta-simulator, which means a design gap exists. It is unlikely that we will have access to all designs in practice. Therefore the meta-simulator will not be able to find the exact design in the target imagery, but a good meta-simulator should output a design that matches the target imagery the most. Similarly, we do four repeated trials, in which each target dataset contains 900 images generated from a virtual city with a pair of unseen $d_{flat}$ and $d_{slope}$ (as shown in the right column in Fig. \ref{fig:exmaples_textures}). 

\textbf{E3: Real-world Target Data}

The third scenario with real-world target imagery is the most challenging but is realistic. Both the design and appearance gap exist between the meta-simulator and the real-world target imagery. We expect the meta-simulator to output a design that matches the target imagery the most. The third target dataset is the Inria dataset with 3600 real-world images per city. We assume that the ground truth of the Inria data are not accessible to the meta-simulation models and the downstream building segmentation models in the training stage. The labels are used only for evaluating the overall performance of these models.
    
\textbf{E4: Computation Time Comparisons.}

The computation time of the NAMS and the MS2 methods is evaluated on our standard hardware, an Intel(R) Core(TM) i7-7700HQ CPU@2.80 GHz with an NVIDIA RTX 2080Ti GPU. Note that, for the MS2 method, we cannot afford the computation cost with the same training process in \cite{Devaranjan2020}, and hence, we estimate the computation cost of the simulation process, based on the number of required simulations and the computation cost of simulating a single image. 

\subsection{Meta-Simulation Training and Inference}

For all meta-simulation models, we use a pre-trained ViT-L16 \cite{Dosovitskiy2020} model to extract 1000-D features $x^{T}$, $x^{S}$, and $x^{G}$ from images. For fair comparisons, we do the same augmentation and feature averaging strategy used in the NAMS method (averaging the features of 72 augmented images with the same design, as discussed in \ref{sec:nams_description}) for all of the other meta-simulation methods. 

\textbf{NAMS.}  In the NAMS model, the encoder $E$, decoder $D$ and predictor $P$ networks are composed of fully-connected layers. The encoder $E$ takes a 84-D vector (a concatenation of $d_{flat} \in [0,1]^{40}$ and $d_{slope} \in [0,1]^{44}$) as input, and it consists of two fully connected layers with 1024 and 2048 hidden units respectively, each followed by a batch normalization layer, a leaky ReLU layer with 0.2 as the slope and a dropout layer with 0.5 as the dropout rate. We use a 10-D vector as the continuous design embedding, $\boldsymbol{z}$, to support the richer design space. Hence, another fully connected layer with 20 hidden units is used as the output layer, with 10 of the output dimensions as $\boldsymbol{\mu}$ and the other 10 as $\boldsymbol{\sigma}$ for sampling the continuous embedding $\boldsymbol{z}$.

We use the same structure in the decoder. The only difference is that we use a 10-D vector as the input $\boldsymbol{z}$ and a 84-D vector as the output data $\boldsymbol{d}$, and a sigmoid function is used in the output layer. Binary cross-entropy loss is used for the reconstruction loss of $\boldsymbol{d}$ in Eq. (\ref{eq:nams_decoder_loss}). We use the same structure in the predictor model, with a 10-D vector as input $\boldsymbol{z}$, and a 1000-D vector as output features. 

We train the NAMS model using a synthetic dataset $(X^{G},Y^{G})$. To form this training dataset, we uniformly sample 1700 pairs of $d_{flat}$ and $d_{slope}$ values and created 9 synthetic images per pair, resulting in 15,000 synthetic images. We train the NAMS model using $(X^{G},Y^{G})$ dataset only once, and apply the same trained model to the target datasets in E1 to E3 for design parameter inference. Additional model training details can be found in the Appendix for each experiment.

\textbf{DR.}  In the DR model, a network consisted of 5 fully-connected hidden layers ($500$ hidden units each layer followed by batch normalization and ReLu layers) is used to predict the 84-D design vector from the 1000-D input features. It is trained with regularizer strength of 0.5, learning rate of 1e-3, batch size of 1000 and train/test split of 80/20, using the same training dataset $(X^{G},Y^{G})$ as is used in NAMS training. The model architecture and the training hyper-parameters are selected by brute-force grid searching.

\textbf{MS2.} The MS2 method does not separate the training and inference stage. It searches design parameters in each target dataset from scratch. The MS2 searching requires 100000 synthetic images for each target dataset (approximated based on \cite{Devaranjan2020}), which will cost 1.8 days with this slow simulator. To search all the 13 target datasets in our experiments from E1 to E3, MS2 will cost unaffordable 23.4 days, let alone the extra searching required for hyper-parameter tuning and the extra needs of (8 times more) images for feature averaging. Hence, we run the simulator for about one month and generate a pool of 464640 synthetic images with 264 images for each design, which is much larger than the training dataset used for DR and NAMS. The synthetic images used in MS2 inference are generated from the pool with replacement. We used the same model architecture and training strategies described in \cite{Devaranjan2020} for the best performance. We only change the input and output data dimensions and adjust the learning rate, to make sure that the training converges at 20 epochs similar to \cite{Devaranjan2020} in our experiments.

\subsection{Task Models and Training Details}

The parameter values inferred by the meta-simulation methods are used in the simulator to generate the optimized synthetic data $(X^{G*},Y^{G*})$ for augmenting the training of the downstream task. 

We then train two contemporary networks for segmentation of satellite imagery, the U-Net \cite{Iglovikov2018} and DeepLabV3 \cite{Chen2017}, which represent popular versions of general architectures for segmentation: the encoder-decoder structure and the feature pyramid structure, respectively. For the best performance, following \cite{Kong2020}, we use a pre-trained Resnet-50 encoder for both of the models, and we train networks for 80,000 mini-batch iterations with a batch size of 7 and the learning rates of 5e-5 and 1e-4 for the DeepLabV3 and U-Net models, respectively. For both networks we drop the learning rate by one order of magnitude after 50,000 iterations of training.

\subsection{Scoring Metrics}

We measure the performance of three meta-simulation methods with two kinds of metrics. 

\textbf{Accuracy of Design Inference.} In the experiments E1 and E2, we know the true design of the target domain imagery. Hence, if we rank-order the incorrect textures according to their similarity to the true texture, we can evaluate whether the meta-simulation method recovers one of the nearest $n_{top}$ textures. For these experiments we use a simple measure of similarity between two image patches \cite{Belongie1998} in HSV color space, to rank nearest textures with $n_{top}=1$ and $n_{top}=10$. Note that, in E1, the target texture is accessible to the simulator, hence, the nearest $n_{top}=1$ texture is exactly the target texture. In the experiments E3 with real-world images, we visually compare the selected synthetic imagery and the real-world target imagery.  

\textbf{Performance of a Task Model.} We then examined the impact of the synthetic imagery augmentation on training downstream task models. We trained two building segmentation models, U-Net and DeepLabV3, with six augmentation strategies for each experiment from E1 to E3 for comparisons: (i) real satellite images only with no augmentation, (ii) augmentation with synthetic images from uniform sampling, (iii) augmentation with synthetic images selected by DR, (iv) augmentation with MS2, (v) augmentation with NAMS; and (vi) augmentation with the ground truth target domain images. We then measure the performance of the segmentation models using the Intersection-over-Union (IoU) metric based on the ground truth labels.

\setlength{\tabcolsep}{4pt}
    \begin{table*}[!t]
        \begin{center}
        \caption{E1 results: accuracy of design inference. The average rates of correctly selecting one of the $n_{top}$ nearest flat-roofed/sloped-roof textures to the target textures among 4 repeated trails in E1 are shown. "Uniform" stands for the uniform sampling method.}
        \label{table:Synthetic_result}
            \begin{tabular}{cccccccccc}
                \hline\noalign{\smallskip}
                & \multicolumn{4}{c}{Top-1} & & \multicolumn{4}{c}{Top-10}\\ \cline{2-5} \cline{7-10}  
                & Uniform  & DR & MS2 & NAMS(ours) & & Uniform & DR & MS2 & NAMS(ours) \\ \cline{2-5} \cline{7-10} 
                Flat-roofed   & 0.025 & \textbf{0.963} & 0.008 & 0.705 & & 0.250 & \textbf{0.983} & 0.418 & 0.893 \\
                Sloped-roof & 0.023 & \textbf{0.888} & 0.008 & 0.323 & & 0.230 & \textbf{0.908} & 0.604 & 0.653 \\
                \hline
            \end{tabular}
        \end{center}
    \end{table*}
\setlength{\tabcolsep}{4pt}
    \begin{table}
        \begin{center}
        \caption{E1 results: performance of a task model. We report the IoU performance of the U-Net and DeepLabV3 models on the target dataset, trained with 6 augmentation strategies. "No" stands for the training with no augmentation. "Uniform" stands for the augmentation with uniformly sampled synthetic images. "GT" stands for the augmentation with the ground truth target domain images.}
        \label{table:task_performance_synthetic}
            \begin{tabular}{ccccccc}
                \hline\noalign{\smallskip}
                Model & No & Uniform & DR & MS2 & NAMS(ours) & GT \\
                \hline
                U-Net     & 47.8 & 77.6 & \textbf{83.4} & 79.1 &  81.8 & 84.3 \\
                DeepLabV3 & 34.8 & 80.5 & \textbf{85.8} & 81.0 &  85.2 & 86.7 \\
                \hline
            \end{tabular}
        \end{center}
    \end{table}

\section{Experimental Results} 
\label{sec:experimental_results}

In this section we present the results of the experiments described in Sec \ref{sec:experimental_design}, and summarized in Table \ref{table:experimental_design_outline}.  We also report the estimated computation time of all of the methods we consider. 

\subsection{E1: Synthetic Target Data with No Design Gap } 
\label{subsec:experimental_results_e1}

\setlength{\tabcolsep}{4pt}
    \begin{table*}[!t]
        \begin{center}
        \caption{E2 results: accuracy of design inference. The average rates of correctly selecting one of the $n_{top}$ nearest flat-roofed/sloped-roof textures to the target textures among 4 repeated trails in E2 are shown. }
        \label{table:Synthetic_result_out}
            \begin{tabular}{cccccccccc}
                \hline\noalign{\smallskip}
                & \multicolumn{4}{c}{Top-1} & & \multicolumn{4}{c}{Top-10}\\ \cline{2-5} \cline{7-10}  
                & Uniform  & DR & MS2 & NAMS(ours) & & Uniform & DR & MS2 & NAMS(ours) \\ \cline{2-5} \cline{7-10} 
                Flat   & 0.025 & \textbf{0.098} & 0.032 & 0.090 & & 0.250 & \textbf{0.890} & 0.328 & 0.708 \\
                Sloped & 0.023 & 0.013 & 0.002 & \textbf{0.105} & & 0.230 & \textbf{0.818} & 0.110 & 0.623 \\
                \hline
            \end{tabular}
        \end{center}
    \end{table*}

\setlength{\tabcolsep}{4pt}
    \begin{table}
        \begin{center}
        \caption{E2 results: performance of a task model. We report the IoU performance of the U-Net and DeepLabV3 models on the target dataset, trained with 6 augmentation strategies.}
        \label{table:task_performance_synthetic_out}
            \begin{tabular}{ccccccc}
                \hline\noalign{\smallskip}
                Model & No & Uniform & DR & MS2 & NAMS(ours) & GT \\
                \hline
                U-Net     & 47.0 & 76.0 & 77.7 & 76.3 &  \textbf{78.7} & 83.0 \\
                DeepLabV3 & 34.5 & 78.6 & 80.7 & 78.5 &  \textbf{81.2} & 85.6 \\
                \hline
            \end{tabular}
        \end{center}
    \end{table}

The accuracy of design inference with our NAMS method in E1 is presented Table \ref{table:Synthetic_result}, in the first column (Top-1) where we report the proportion of target images for which NAMS recovers the true parameters, compared with (i) uniform sampling, (ii) searching with MS2 and (iii) regressing with DR. The results in Table \ref{table:Synthetic_result} shown that DR still performs well for this in-domain test with no design gaps, and NAMS is substantially more accurate than MS2, even with a much smaller training dataset. In the second column (Top-10) in Table \ref{table:Synthetic_result}, we see that NAMS achieves much greater accuracy when evaluating whether it finds one of the top-10 best patches (by our measure), suggesting that it is still finds a more appropriate design, even if it is not the best design.

The downstream task performance results of this experiment are presented in Table \ref{table:task_performance_synthetic}, for both the U-Net and DeepLabV3 models. The results indicate that DR performs pretty well as expected. NAMS improves average performance by 5.8\% and 6.1\% on average for the U-Net and DeepLabV3 models, which is better than MS2, and closes most of the performance gap between a uniform sampling and the best possible performance, achieved when design parameters perfectly match the testing synthetic dataset.

\subsection{E2: Synthetic Target Data with a Design Gap } 
\label{subsec:experimental_results_e2}

Here, it is impossible to check if the meta-simulation methods inferred the correct design parameters, since the target value is not in the training space. However, we can still rank the nearest $n_{top}$ textures according to their similarity to the true texture in HSV color space, similar to the E1 experiments. As shown in Table \ref{table:Synthetic_result_out}, all of the results get smaller compared to the E1 tests in Table \ref{table:Synthetic_result}, and NAMS can still outperform MS2 in all cases. One interesting thing to note is that, NAMS is able to achieve better accuracy then DR for searching the top-1 sloped roof texture, suggesting that NAMS is relatively more robust to infer out-of-domain targets than DR.

Next, as shown in Table \ref{table:task_performance_synthetic_out}, NAMS improves average performance for U-Net by 3.8\% and DeepLabV3 by 3.5\% which is much better than MS2, while the improvement of DR is only 2.5\% for U-Net and 2.8\% for DeepLabV3. It indicated that, the robustness of the NAMS method is better than DR when design gap exists between the training and target dataset.

\subsection{E3: Real-world Target Data} 
\label{subsec:experimental_results_e3}

\begin{figure*}[!t]
    \centering
    \includegraphics[height=5.5cm]{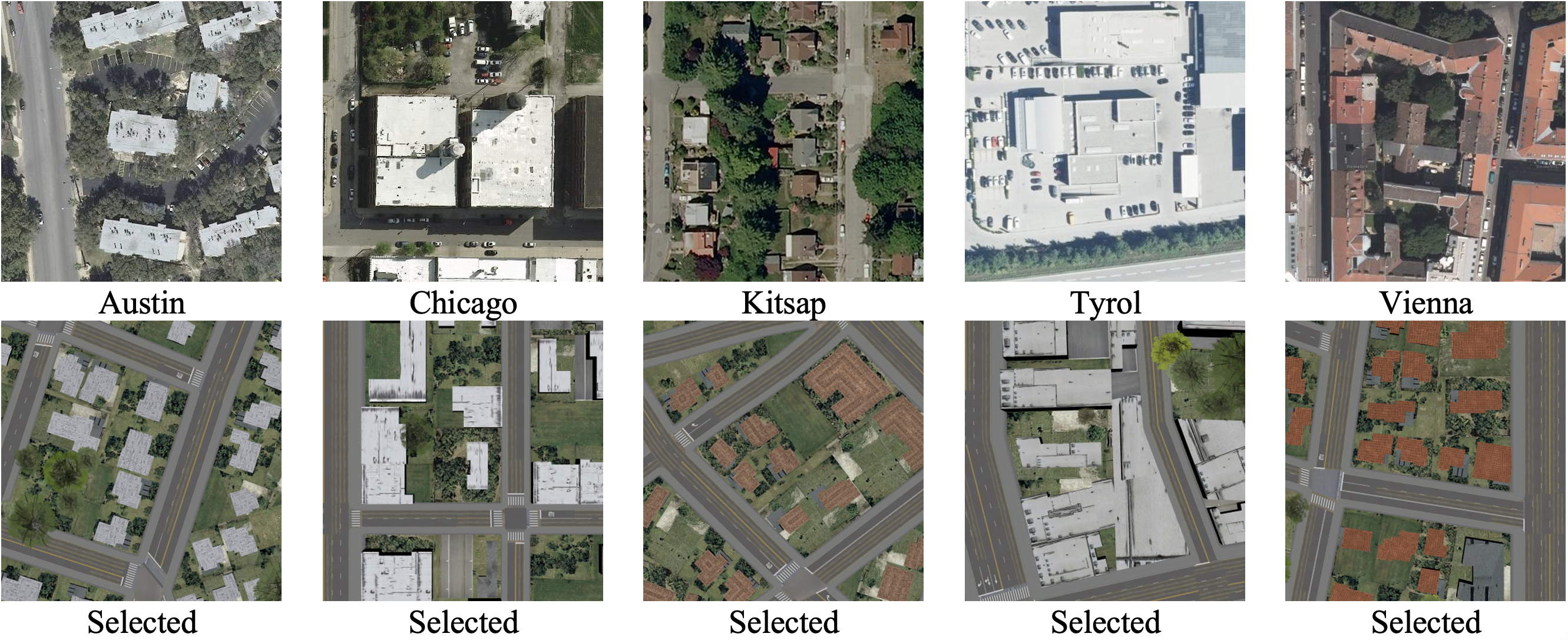}
    \caption{E3 results: accuracy of design inference. Examples of the target images and the selected images by NAMS are shown.}
    \label{fig:Meta_sim_result}
\end{figure*}

\setlength{\tabcolsep}{4pt}
    \begin{table}
        \begin{center}
        \caption{E3 results: performance of a task model. We report the IoU performance of the U-Net and DeepLabV3 models on the target dataset, trained with 6 augmentation strategies.}
        \label{table:task_performance_real}
            \begin{tabular}{ccccccc}
                \hline\noalign{\smallskip}
                Model & No & Uniform & DR & MS2 & NAMS(ours) & GT \\
                \hline
                U-Net     & 51.7 & 53.4 & 53.3 & 52.7 &  \textbf{54.0} & 58.8 \\
                DeepLabV3 & 46.9 & 49.8 & 50.2 & 50.4 &  \textbf{50.6} & 58.9 \\
                \hline
            \end{tabular}
        \end{center}
    \end{table}

In E3 experiment, we evaluate our method for building segmentation on real satellite imagery. Our goal is again to choose $d_{flat}$ and $d_{slope}$, the flat and sloped rooftop textures that best match the rooftops in the real-world target imagery.

The results of E3 are presented in Table \ref{table:task_performance_real}, and show that NAMS outperforms both the MS2 and the DR approach, on average, across the five testing cities, and the improvement from uniform sampling is 1.2\% for U-Net and 1.5\% for DeepLabV3. In contrast to the synthetic experiments E1 and E2, however, the benefits associated with NAMS are generally lower. We hypothesize that this is likely due to the substantial sim-to-real appearance gap between our synthetic imagery and the real-world imagery. Fig. \ref{fig:Meta_sim_result} presents examples of the real-world target imagery, and the synthetic designs that were chosen by NAMS for those cities. The designs chosen are qualitatively similar to those in the target cities, however, the sim-to-real gap is apparent in the imagery.

\subsection{E4: Computation Time Comparison}
\label{sec:experiment_computation_time}

Here we present the computation time of NAMS on our benchmark tasks, and compare it with the computation time of the non-amortized method, MS2. Key metrics of each approach are summarized in Table \ref{table:Computation_time}. NAMS must be trained on a collection of simulations first. After this training however, it can infer design parameters on new target domains (or individual images) without additional simulation. By contrast, MS2 requires no initial simulation or training, but it requires a large number of simulations for each target domain, and this process must be repeated from scratch for each new target domain.

We examine the overhead imagery task for example, which motivated the development of NAMS. NAMS required 15,000 initial simulations and 13h of training, whereas MS2 requires no offline training or simulations. However, NAMS requires 0.7h to infer parameters for each new target domain, whereas MS2 requires 1.8 days (16.2 days if average strategy is applied) with the same training process in \cite{Devaranjan2020}. Therefore NAMS scales much more efficiently with the number of target domains. Furthermore, NAMS can infer parameters for a small group of 9 images. In this estimation we assumed that each target domain comprises 5000 images, following \cite{Devaranjan2020}, \cite{Kar2019a}. However, if we are satisfied with parameter estimates obtained on just 50 target images, in which case NAMS requires only 25s per domain (0.5 sec/image $*$ 50 images).

\setlength{\tabcolsep}{4pt}
\begin{table}
    \begin{center}
        \caption{Key computation time metrics for NAMS and MS2 on our benchmark task. $N_d$ denotes the number of target domains. Each target domain is assumed to comprise 5000 images.}
        \label{table:Computation_time}
        \begin{tabular}{clc}
        \hline
            & Experiments           & Overhead imagery  \\ 
            \cline{2-3}
            & Simulation time       & 1.6s/image    \\ 
            \hline
            \parbox[t]{2mm}{\multirow{ 4}{*}{\rotatebox[origin=c]{90}{NAMS}}}
            & Total simulations required        & 15,000   \\
            & Training time                     & 13h \\
            & Inference per image               & 0.5s\\
            & Inference per target domain       & 0.7h      \\
            \hline
            \parbox[t]{2mm}{\multirow{4}{*}{\rotatebox[origin=c]{90}{MS2}}}    
            & Total simulations          & 100,000*$N_d$     \\
            & Training time               & 0    \\
            & Inference per image       & N/A  \\
            & Inference per target domain    & 1.8 days (estimated)\\ 
            \hline
        \end{tabular}
    \end{center}
\end{table}

\begin{figure*}[!t]
    \centering
    \begin{tabular}{c}
        \includegraphics[height=4cm]{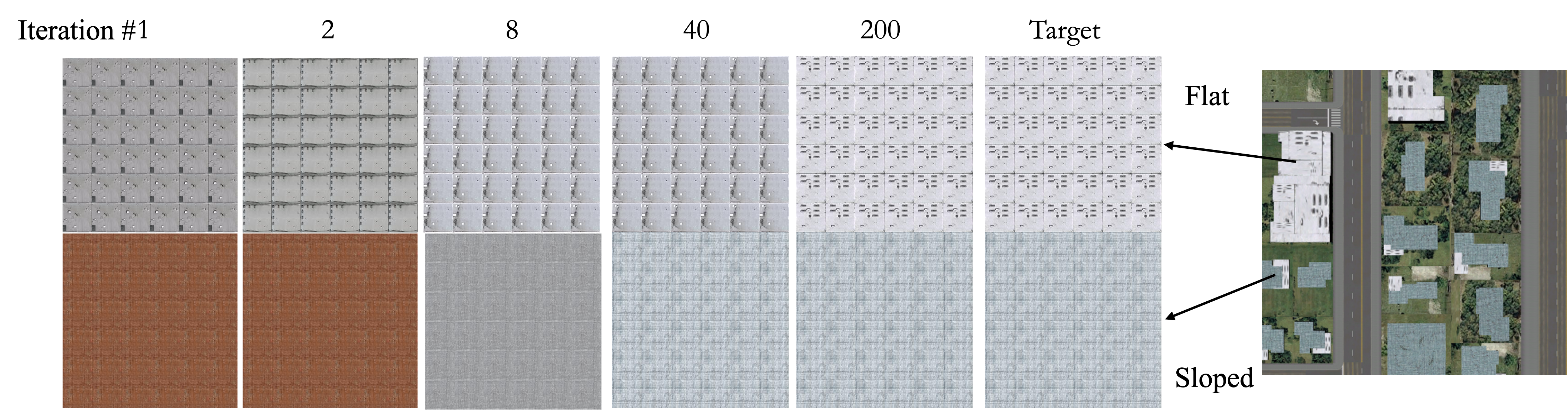} \\
        \small (a)
    \end{tabular}

    \begin{tabular}{c}
        \includegraphics[height=4cm]{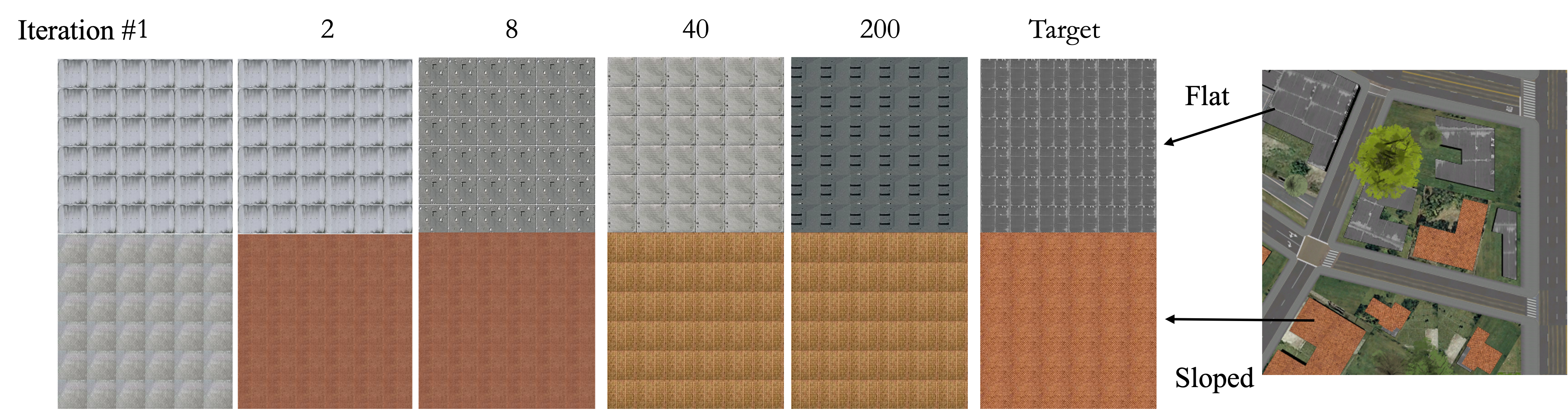} \\
        \small (b)
    \end{tabular}

    \begin{tabular}{c}
        \includegraphics[height=4cm]{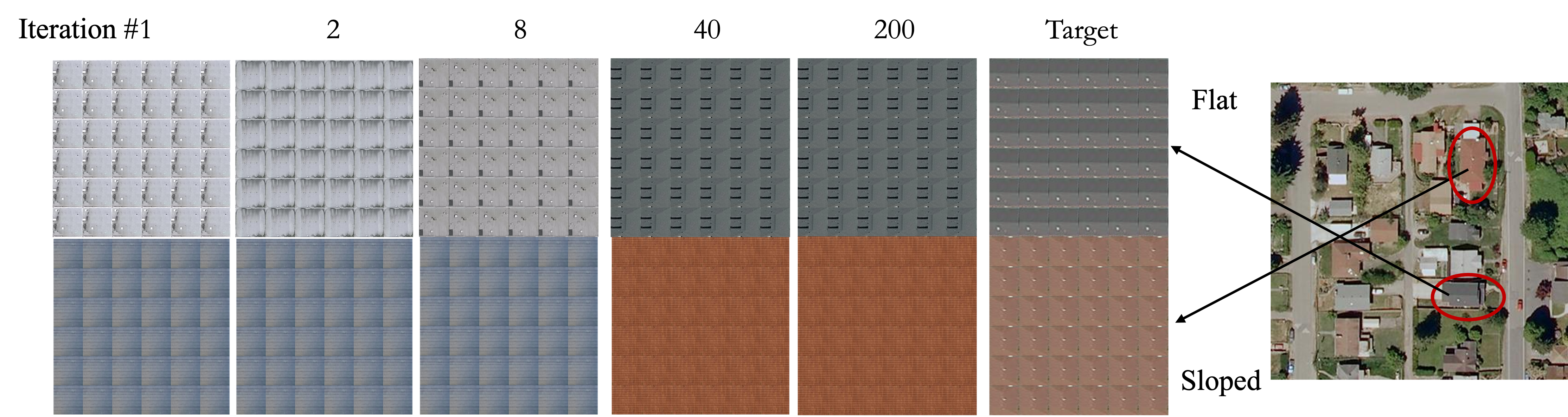} \\
        \small (c)
    \end{tabular}
    
    \caption{The searching process for the NAMS method in experiments E1 to E3 is shown compared with the target design and image.}
    \label{fig:iter}
\end{figure*}

\section{Conclusions}\label{sec:conclusion}
In this paper, we first investigate and compare existing meta-simulation methods for overhead imagery. We evaluate two state-of-the-art methods, direct regression and MS2, on several carefully-designed experiments. A novel NAMS method is then proposed to address the unique challenges associated with overhead imagery. NAMS is composed of two key ideas. First, we propose a continuous embedding for discrete/categorical design parameters, by using a VAE model. Second, we learn a Neural-Adjoint network for inferring image data from design space to by-pass the black-box simulator. We show in several synthetic and real-world experiments that, by using NAMS, synthetic images with the target design can be rapidly and accurately generated for the training of downstream task algorithms. Compared to existing methods, NAMS is much more robust to sim-to-real gaps and computationally efficient as the number of unique target domains grows larger.

\appendix
\begin{figure}
    \centering
    \includegraphics[height=3.5cm]{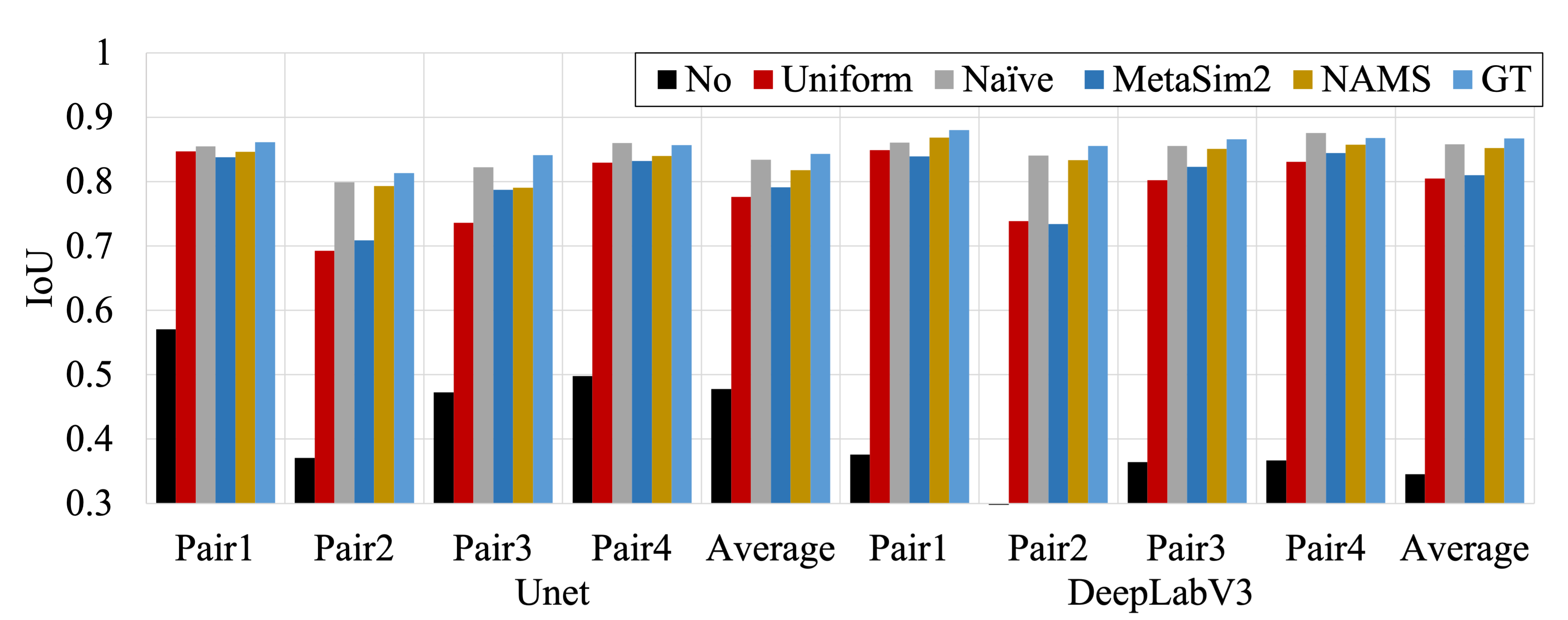}
    \caption{City-wise performance comparisons of two models (a) U-Net and (b) DeepLabV3 in experiment E1 are shown.}
    \label{fig:city_in_domain}
\end{figure}

\section*{Appendix for E1}

\textbf{NAMS training and design parameter inference}. We set $\lambda_D=1$, $\lambda_{KLD}=0.0005$ and $\lambda_P=0.005$ in the loss function in Eq. \ref{eq:nams_overall_loss}, and the model is trained using ADAM optimizer for 50000 epochs. The batch size is set to 64. We set $r=0$ for design parameter inference in this synthetic image experiment, since there
is no need to diversify the simulated training data for such a
simple experiment with no diversity in the target design. To infer designs, we use $M=1000$ initialization points and $S=50$. For each initialization we performed gradient
descent with the ADAM optimizer ($l=0.01$, $\beta_1=0.9$, $\beta_2=0.999$) for 200 iterations.

\textbf{The searching process}. We show the searching process of one target image in Fig. \ref{fig:iter} (a). The selected designs, i.e., the flat and sloped roof textures, at different iterations are shown, for comparisons with the target textures.  

\textbf{City-wise result comparisons}. We do 4 repeated trials in 4 synthetic cities, each with a randomly-selected pair of flat and sloped roof textures in the target domain. We presented average performance of our methods in the main manuscript; in Fig. \ref{fig:city_in_domain} we present the city-wise performance. The proposed NAMS outperforms uniform sampling and MS2 on all of the target image pairs, and across two models.

\begin{figure}
    \centering
    \includegraphics[height=3.5cm]{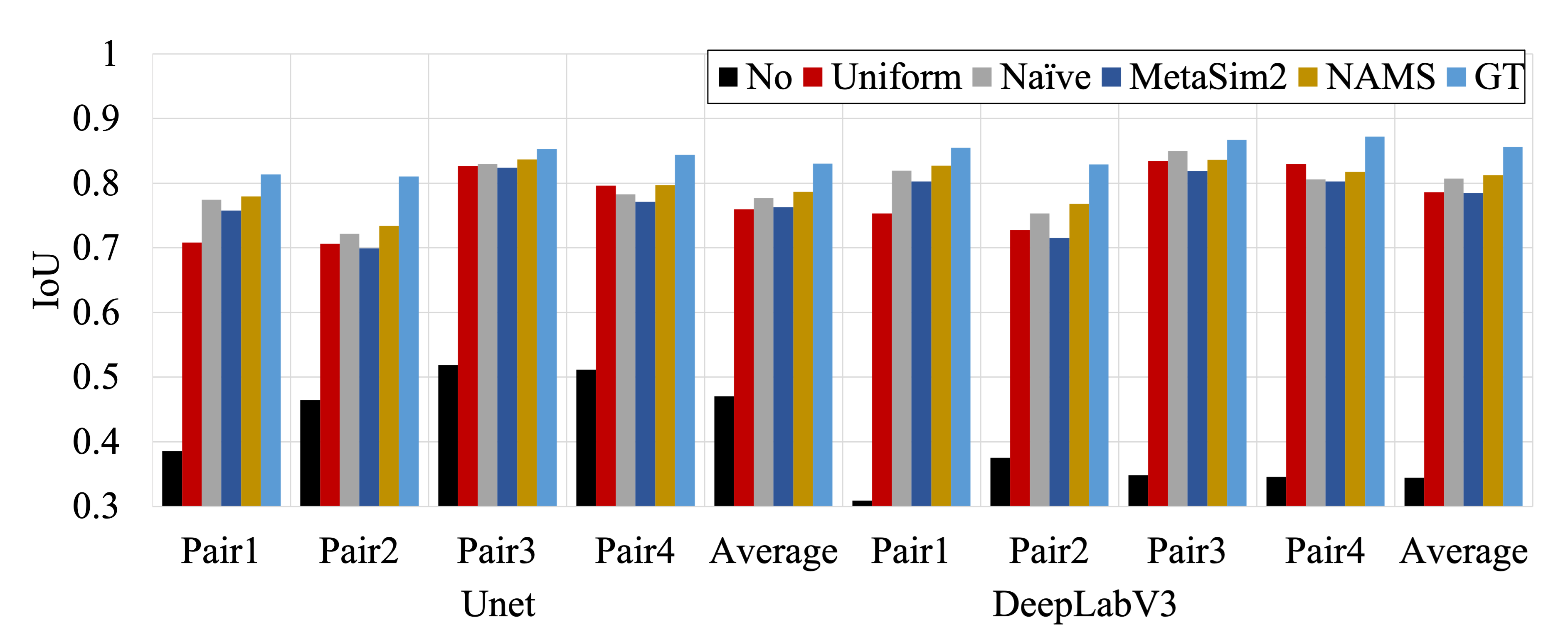}
    \caption{City-wise performance comparisons of two models (a) U-Net and (b) DeepLabV3 in experiment E2 are shown.}
    \label{fig:city_out_domain}
\end{figure}

\section*{Appendix for E2}
\textbf{NAMS training and design parameter inference}. We use the same NAMS model trained in E1, and we use the same settings for design parameter inference in this synthetic image experiment.

\textbf{The searching process}. We show the searching process of one target image in Fig.  \ref{fig:iter} (b). Note that, the target textures are not in the training domain, and hence, the aim of searching is to find the closest textures compared to the target textures. 

\textbf{City-wise result comparisons}. We also generated 4 synthetic cities, each with new flat and sloped roof textures. In Fig. \ref{fig:city_out_domain} we present the city-wise performance. The proposed NAMS outperforms uniform sampling and DR in 7 out of 8 cases and outperforms MS2 in all cases.

\begin{figure}
    \centering
    \includegraphics[height=3.4cm]{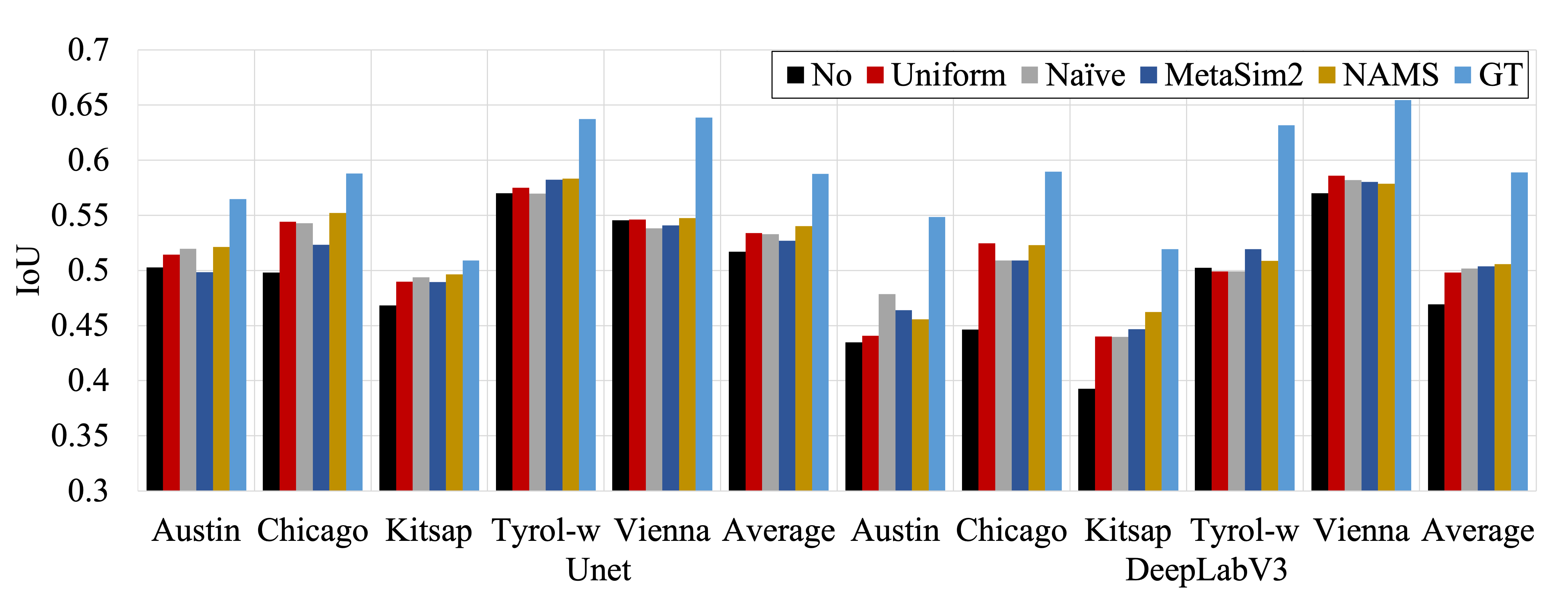}
    \caption{City-wise performance comparisons of two models (a) U-Net and (b) DeepLabV3 in experiment E3 are shown.}
    \label{fig:city_real}
\end{figure}

\section*{Appendix for E3}
\textbf{NAMS training and design parameter inference}. We use the same NAMS model trained in E1. To infer designs, we set $r=0.25$, for the reason mentioned in Section \ref{sec:nams_description}, and we used the same method and other hyper-parameters as we used in the synthetic overhead experiments.

\textbf{The searching process}. We show the searching process of one real target image in Fig.  \ref{fig:iter} (c). Note that, the target textures are also not in the training domain, and hence, the aim of searching is to find the closest textures in the training domain. 

\textbf{City-wise result comparisons}. We examined the performance of meta-simulation on five real-world test cities of satellite imagery. In Fig. \ref{fig:city_out_domain} we present the city-wise performance. The proposed NAMS outperforms uniform sampling in 7/8 cases, outperforms DR in 6/8 cases, and outperforms MS2 in 5/6 cases.

\bibliographystyle{unsrt}  
\bibliography{references}

\begin{thebibliography}{10}

\bibitem{Iglovikov2018}
Vladimir~I. Iglovikov, Selim Seferbekov, Alexander~V. Buslaev, and Alexey
  Shvets.
\newblock {TernausNetV2: Fully Convolutional Network for Instance
  Segmentation}.
\newblock In {\em Computer Vision and Pattern Recognition (CVPR)}, 2018.

\bibitem{Demir2018}
Ilke Demir, Krzysztof Koperski, David Lindenbaum, Guan Pang, Jing Huang, Saikat
  Basu, Forest Hughes, Devis Tuia, and Ramesh Raskar.
\newblock {DeepGlobe 2018: A Challenge to Parse the Earth through Satellite
  Images}.
\newblock In {\em Computer Vision and Pattern Recognition (CVPR)}, pages
  172--181, 2018.

\bibitem{Maggiori2017}
Emmanuel Maggiori, Yuliya Tarabalka, Guillaume Charpiat, Pierre Alliez,
  Emmanuel Maggiori, Yuliya Tarabalka, Guillaume Charpiat, Pierre Alliez, and
  Can Semantic.
\newblock {Can Semantic Labeling Methods Generalize to Any City ? The Inria
  Aerial Image Labeling Benchmark}.
\newblock In {\em International Geoscience and Remote Sensing Symposium
  (IGARSS)}, pages 3226--3229, 2017.

\bibitem{Wang2017}
Yida Wang, David~Joseph Tan, Nassir Navab, and Federico Tombari.
\newblock {Semantic scene completion from a single depth image}.
\newblock {\em Computer vision and pattern (CVPR)}, pages 426--434, 2017.

\bibitem{Huang2018}
Bohao Huang, Kangkang Lu, Nicolas Audebert, Andrew Khalel, Yuliya Tarabalka,
  Jordan~M. Malof, Alexandre Boulch, Bertrand~Le Saux, Leslie Collins, Kyle
  Bradbury, Sebastien Lefevre, and Motaz El-Saban.
\newblock {Large-scale semantic classification: outcome of the first year of
  inria aerial image labeling benchmark}.
\newblock In {\em International Geoscience and Remote Sensing Symposium}, 2018.

\bibitem{Kong2020}
Fanjie Kong, Bohao Huang, Kyle Bradbury, and J.M. Jordan~M Malof.
\newblock {The Synthinel-1 dataset : a collection of high resolution synthetic
  overhead imagery for building segmentation}.
\newblock In {\em Workshop on Advanced Computer Vision (WACV)}, page~1, 2020.

\bibitem{Ward2018}
Chris~M Ward, Josh Harguess, and Cameron Hilton.
\newblock Ship classification from overhead imagery using synthetic data and
  domain adaptation.
\newblock In {\em OCEANS 2018 MTS/IEEE Charleston}, pages 1--5. IEEE, 2018.

\bibitem{Shermeyer2021}
Jacob Shermeyer, Thomas Hossler, Adam Van~Etten, Daniel Hogan, Ryan Lewis, and
  Daeil Kim.
\newblock Rareplanes: Synthetic data takes flight.
\newblock In {\em Proceedings of the IEEE/CVF Winter Conference on Applications
  of Computer Vision}, pages 207--217, 2021.

\bibitem{Xu}
Yang Xu, Bohao Huang, Xiong Luo, Kyle Bradbury, and Jordan~M Malof.
\newblock {SIMPL : Generating Synthetic Overhead Imagery to Address Zero-shot
  and Few-Shot Detection Problems}.

\bibitem{Hu2021}
Wei Hu, Tyler Feldman, Yanchen~J Ou, Natalie Tarn, Baoyan Ye, Yang Xu, Jordan~M
  Malof, and Kyle Bradbury.
\newblock Wind turbine detection with synthetic overhead imagery.
\newblock In {\em 2021 IEEE International Geoscience and Remote Sensing
  Symposium IGARSS}, pages 4908--4911. IEEE, 2021.

\bibitem{Han2017}
Sanghui Han, Alex Fafard, John Kerekes, Michael Gartley, Emmett Ientilucci,
  Andreas Savakis, Charles Law, Jason Parhan, Matt Turek, Keith Fieldhouse, and
  Todd Rovito.
\newblock {Efficient generation of image chips for training deep learning
  algorithms}.
\newblock 1020203(May 2017):1020203, 2017.

\bibitem{Ros2016}
German Ros, Laura Sellart, Joanna Materzynska, David Vazquez, and Antonio~M.
  Lopez.
\newblock {The SYNTHIA Dataset: A Large Collection of Synthetic Images for
  Semantic Segmentation of Urban Scenes}.
\newblock {\em Proceedings of the IEEE Computer Society Conference on Computer
  Vision and Pattern Recognition}, 2016-Decem(600388):3234--3243, 2016.

\bibitem{Richter2016}
Stephan~R Richter, Vibhav Vineet, Stefan Roth, and Vladlen Koltun.
\newblock {Playing for Data: Ground Truth from Computer Games}.
\newblock In {\em European conference on computer vision}, volume 9908, pages
  102--118, 2016.

\bibitem{Tobin2017}
Josh Tobin, Rachel Fong, Alex Ray, Jonas Schneider, Wojciech Zaremba, and
  Pieter Abbeel.
\newblock {Domain randomization for transferring deep neural networks from
  simulation to the real world}.
\newblock {\em IEEE International Conference on Intelligent Robots and
  Systems}, 2017-Septe:23--30, 2017.

\bibitem{Zhang2017}
Yinda Zhang, Shuran Song, Ersin Yumer, Manolis Savva, Joon~Young Lee, Hailin
  Jin, and Thomas Funkhouser.
\newblock {Physically-based rendering for indoor scene understanding using
  convolutional neural networks}.
\newblock {\em Proceedings - 30th IEEE Conference on Computer Vision and
  Pattern Recognition, CVPR 2017}, 2017-Janua:5057--5065, 2017.

\bibitem{Shafaei2016}
Alireza Shafaei, James~J. Little, and Mark Schmidt.
\newblock {Play and learn: Using video games to train computer vision models}.
\newblock {\em British Machine Vision Conference 2016, BMVC 2016},
  2016-Septe:26.1--26.13, 2016.

\bibitem{Qureshi2008}
Faisal Qureshi and Demetri Terzopoulos.
\newblock {Smart camera networks in virtual reality}.
\newblock {\em Proceedings of the IEEE}, 96(10):1640--1656, 2008.

\bibitem{Taylor2007}
Geoffrey~R. Taylor, Andrew~J. Chosak, and Paul~C. Brewer.
\newblock {OVVV: Using virtual worlds to design and evaluate surveillance
  systems}.
\newblock {\em Proceedings of the IEEE Computer Society Conference on Computer
  Vision and Pattern Recognition}, 2007.

\bibitem{Tremblay2018}
Jonathan Tremblay, Aayush Prakash, David Acuna, Mark Brophy, Varun Jampani, Cem
  Anil, Thang To, Eric Cameracci, Shaad Boochoon, and Stan Birchfield.
\newblock {Training deep networks with synthetic data: Bridging the reality gap
  by domain randomization}.
\newblock {\em IEEE Computer Society Conference on Computer Vision and Pattern
  Recognition Workshops}, 2018-June:1082--1090, 2018.

\bibitem{Sankaranarayanan2018}
Swami Sankaranarayanan, Yogesh Balaji, Arpit Jain, Ser~Nam Lim, and Rama
  Chellappa.
\newblock {Learning from Synthetic Data: Addressing Domain Shift for Semantic
  Segmentation}.
\newblock In {\em Computer Vision and Pattern Recognition}, 2018.

\bibitem{Kar2019a}
Amlan Kar, Aayush Prakash, Ming-Yu Liu, Eric Cameracci, Justin Yuan, Matt
  Rusiniak, David Acuna, Antonio Torralba, and Sanja Fidler.
\newblock Meta-sim: Learning to generate synthetic datasets.
\newblock In {\em Proceedings of the IEEE/CVF International Conference on
  Computer Vision}, pages 4551--4560, 2019.

\bibitem{Devaranjan2020}
Jeevan Devaranjan, Amlan Kar, and Sanja Fidler.
\newblock {Meta-Sim2: Unsupervised Learning of Scene Structure for Synthetic
  Data Generation}.
\newblock {\em Lecture Notes in Computer Science (including subseries Lecture
  Notes in Artificial Intelligence and Lecture Notes in Bioinformatics)}, 12362
  LNCS:715--733, 2020.

\bibitem{hoffman_2018}
Judy Hoffman, Eric Tzeng, Taesung Park, Jun-Yan Zhu, Phillip Isola, Kate
  Saenko, Alexei Efros, and Trevor Darrell.
\newblock Cycada: Cycle-consistent adversarial domain adaptation.
\newblock {\em PMLR}, Jul 2018.

\bibitem{huang2018multimodal}
Xun Huang, Ming-Yu Liu, Serge Belongie, and Jan Kautz.
\newblock Multimodal unsupervised image-to-image translation.
\newblock In {\em Proceedings of the European conference on computer vision
  (ECCV)}, pages 172--189, 2018.

\bibitem{Kulkarni2015Deep}
Tejas~D Kulkarni, William~F. Whitney, Pushmeet Kohli, and Josh Tenenbaum.
\newblock Deep convolutional inverse graphics network.
\newblock {\em Advances in Neural Information Processing Systems}, 28, 2015.

\bibitem{Du2021Auto-Tuned}
Yuqing Du, Olivia Watkins, Trevor Darrell, Pieter Abbeel, and Deepak Pathak.
\newblock Auto-tuned sim-to-real transfer.
\newblock {\em arXiv:2104.07662}, 5 2021.

\bibitem{Mansinghka2013Approximate}
Vikash~K. Mansinghka, Tejas~D. Kulkarni, Yura~N. Perov, and Joshua~B.
  Tenenbaum.
\newblock Approximate bayesian image interpretation using generative
  probabilistic graphics programs.
\newblock {\em arXiv:1307.0060 [cs, stat]}, 6 2013.

\bibitem{Kulkarni2015Picture:}
Tejas~D Kulkarni, Pushmeet Kohli, Joshua~B Tenenbaum, and Vikash Mansinghka.
\newblock Picture: A probabilistic programming language for scene perception.
\newblock pages 4390--4399, Boston, MA, USA, 6 2015. 2015 IEEE Conference on
  Computer Vision and Pattern Recognition (CVPR), IEEE.

\bibitem{Yildirim2015}
Ilker Yildirim and Winrich~A Freiwald.
\newblock Efﬁcient analysis-by-synthesis in vision: A computational
  framework, behavioral tests, and comparison with neural representations.
\newblock {\em Thirty-seventh annual conference of the cognitive science
  society}, page~6, 2015.

\bibitem{Louppe2019Adversarial}
Gilles Louppe, Joeri Hermans, and Kyle Cranmer.
\newblock Adversarial variational optimization of non-differentiable
  simulators.
\newblock pages 1438--1447. The 22nd International Conference on Artificial
  Intelligence and Statistics, PMLR, 4 2019.

\bibitem{Ganin2018Synthesizing}
Yaroslav Ganin, Tejas Kulkarni, Igor Babuschkin, S.~M.~Ali Eslami, and Oriol
  Vinyals.
\newblock Synthesizing programs for images using reinforced adversarial
  learning.
\newblock pages 1666--1675. International Conference on Machine Learning, PMLR,
  7 2018.

\bibitem{Mellor2019Unsupervised}
John F.~J. Mellor, Eunbyung Park, Yaroslav Ganin, Igor Babuschkin, Tejas
  Kulkarni, Dan Rosenbaum, Andy Ballard, Theophane Weber, Oriol Vinyals, and
  S.~M.~Ali Eslami.
\newblock Unsupervised doodling and painting with improved spiral.
\newblock {\em arXiv:1910.01007 [cs, stat]}, 10 2019.

\bibitem{Behl2020AutoSimulate:}
Harkirat~Singh Behl, Atılım~Güneş Baydin, Ran Gal, Philip H.~S. Torr, and
  Vibhav Vineet.
\newblock Autosimulate: (quickly) learning synthetic data generation.
\newblock {\em arXiv:2008.08424}, 8 2020.

\bibitem{Ruiz2019LEARNING}
Nataniel Ruiz, Samuel Schulter, and Manmohan Chandraker.
\newblock Learning to simulate.
\newblock page~12, 2019.

\bibitem{Wrenninge2018}
Magnus Wrenninge and Jonas Unger.
\newblock {Synscapes: A Photorealistic Synthetic Dataset for Street Scene
  Parsing}.
\newblock 2018.

\bibitem{Prakash2019}
Aayush Prakash, Shaad Boochoon, Mark Brophy, David Acuna, Eric Cameracci,
  Gavriel State, Omer Shapira, and Stan Birchfield.
\newblock {Structured domain randomization: Bridging the reality gap by
  context-aware synthetic data}.
\newblock In {\em 2019 International Conference on Robotics and Automation
  (ICRA)}, pages 7249--7255. IEEE, 2019.

\bibitem{Gaidon2016}
Adrien Gaidon, Qiao Wang, Yohann Cabon, and Eleonora Vig.
\newblock {Virtual worlds as proxy for multi-object tracking analysis}.
\newblock In {\em Proceedings of the IEEE conference on computer vision and
  pattern recognition}, volume 2016-Decem, pages 4340--4349, 2016.

\bibitem{Dosovitskiy2017}
Alexey Dosovitskiy, German Ros, Felipe Codevilla, Antonio Lopez, and Vladlen
  Koltun.
\newblock {CARLA: An open urban driving simulator}.
\newblock In {\em Conference on robot learning}, pages 1--16. PMLR, 2017.

\bibitem{Alhaija2018}
Hassan~Abu Alhaija, Siva~Karthik Mustikovela, Lars Mescheder, Andreas Geiger,
  and Carsten Rother.
\newblock {Augmented reality meets computer vision: Efficient data generation
  for urban driving scenes}.
\newblock {\em International Journal of Computer Vision}, 126(9):961--972,
  2018.

\bibitem{Armeni2019}
Iro Armeni, Zhi-Yang He, JunYoung Gwak, Amir~R Zamir, Martin Fischer, Jitendra
  Malik, and Silvio Savarese.
\newblock {3d scene graph: A structure for unified semantics, 3d space, and
  camera}.
\newblock In {\em Proceedings of the IEEE/CVF International Conference on
  Computer Vision}, pages 5664--5673, 2019.

\bibitem{Handa2016}
Ankur Handa, Viorica Patraucean, Vijay Badrinarayanan, Simon Stent, and Roberto
  Cipolla.
\newblock {Understanding real world indoor scenes with synthetic data}.
\newblock In {\em Proceedings of the IEEE conference on computer vision and
  pattern recognition}, pages 4077--4085, 2016.

\bibitem{McCormac2017}
John McCormac, Ankur Handa, Stefan Leutenegger, and Andrew~J. Davison.
\newblock {SceneNet RGB-D: Can 5M Synthetic Images Beat Generic ImageNet
  Pre-training on Indoor Segmentation?}
\newblock {\em Proceedings of the IEEE International Conference on Computer
  Vision}, 2017-October:2697--2706, 2017.

\bibitem{Savva2019}
Manolis Savva, Abhishek Kadian, Oleksandr Maksymets, Yili Zhao, Erik Wijmans,
  Bhavana Jain, Julian Straub, Jia Liu, Vladlen Koltun, Jitendra Malik, Devi
  Parikh, and Dhruv Batra.
\newblock {Habitat: A platform for embodied AI research}.
\newblock {\em Proceedings of the IEEE International Conference on Computer
  Vision}, 2019-October:9338--9346, 2019.

\bibitem{Wu2018}
Yi~Wu, Yuxin Wu, Georgia Gkioxari, and Yuandong Tian.
\newblock {Building generalizable agents with a realistic and rich 3D
  environment}.
\newblock {\em 6th International Conference on Learning Representations, ICLR
  2018 - Workshop Track Proceedings}, pages 1--15, 2018.

\bibitem{Tassa2018}
Yuval Tassa, Yotam Doron, Alistair Muldal, Tom Erez, Yazhe Li, Diego de~Las
  Casas, David Budden, Abbas Abdolmaleki, Josh Merel, Andrew Lefrancq, Timothy
  Lillicrap, and Martin Riedmiller.
\newblock {DeepMind Control Suite}.
\newblock 2018.

\bibitem{Todorov2012}
Emanuel Todorov, Tom Erez, and Yuval Tassa.
\newblock {Mujoco: A physics engine for model-based control}.
\newblock In {\em 2012 IEEE/RSJ International Conference on Intelligent Robots
  and Systems}, pages 5026--5033. IEEE, 2012.

\bibitem{Sadeghi2016}
Fereshteh Sadeghi and Sergey Levine.
\newblock {Cad2rl: Real single-image flight without a single real image}.
\newblock {\em arXiv preprint arXiv:1611.04201}, 2016.

\bibitem{Brockman2016}
Greg Brockman, Vicki Cheung, Ludwig Pettersson, Jonas Schneider, John Schulman,
  Jie Tang, and Wojciech Zaremba.
\newblock {Openai gym}.
\newblock {\em arXiv preprint arXiv:1606.01540}, 2016.

\bibitem{Butler2012}
Daniel~J Butler, Jonas Wulff, Garrett~B Stanley, and Michael~J Black.
\newblock {A naturalistic open source movie for optical flow evaluation}.
\newblock In {\em European conference on computer vision}, pages 611--625.
  Springer, 2012.

\bibitem{Shugrina2019}
Maria Shugrina, Ziheng Liang, Amlan Kar, Jiaman Li, Angad Singh, Karan Singh,
  and Sanja Fidler.
\newblock {Creative flow+ dataset}.
\newblock {\em Proceedings of the IEEE Computer Society Conference on Computer
  Vision and Pattern Recognition}, 2019-June:5379--5388, 2019.

\bibitem{Puig2018}
Xavier Puig, Kevin Ra, Marko Boben, Jiaman Li, Tingwu Wang, Sanja Fidler, and
  Antonio Torralba.
\newblock {VirtualHome: Simulating Household Activities Via Programs}.
\newblock {\em Proceedings of the IEEE Computer Society Conference on Computer
  Vision and Pattern Recognition}, pages 8494--8502, 2018.

\bibitem{Kolve2017}
Eric Kolve, Roozbeh Mottaghi, Winson Han, Eli VanderBilt, Luca Weihs, Alvaro
  Herrasti, Daniel Gordon, Yuke Zhu, Abhinav Gupta, and Ali Farhadi.
\newblock {Ai2-thor: An interactive 3d environment for visual ai}.
\newblock {\em arXiv preprint arXiv:1712.05474}, 2017.

\bibitem{Gao2019}
Xiaofeng Gao, Ran Gong, Tianmin Shu, Xu~Xie, Shu Wang, and Song-Chun Zhu.
\newblock {Vrkitchen: an interactive 3d virtual environment for task-oriented
  learning}.
\newblock {\em arXiv preprint arXiv:1903.05757}, 2019.

\bibitem{Qureshi2007}
Faisal~Z. Qureshi and Demetri Terzopoulos.
\newblock {Surveillance in virtual reality: System design and multi-camera
  control}.
\newblock {\em Proceedings of the IEEE Computer Society Conference on Computer
  Vision and Pattern Recognition}, 2007.

\bibitem{Ren2020}
Simiao Ren, Willie Padilla, and Jordan Malof.
\newblock Benchmarking deep inverse models over time, and the neural-adjoint
  method.
\newblock In H.~Larochelle, M.~Ranzato, R.~Hadsell, M.~F. Balcan, and H.~Lin,
  editors, {\em Advances in Neural Information Processing Systems}, volume~33,
  pages 38--48. Curran Associates, Inc., 2020.

\bibitem{peurifoy2018nanophotonic}
John Peurifoy, Yichen Shen, Li~Jing, Yi~Yang, Fidel Cano-Renteria, Brendan~G
  DeLacy, John~D Joannopoulos, Max Tegmark, and Marin Solja{\v{c}}i{\'c}.
\newblock Nanophotonic particle simulation and inverse design using artificial
  neural networks.
\newblock {\em Science advances}, 4(6):eaar4206, 2018.

\bibitem{Gomez-Bombarelli2018}
Rafael G{\'{o}}mez-Bombarelli, Jennifer~N. Wei, David Duvenaud,
  Jos{\'{e}}~Miguel Hern{\'{a}}ndez-Lobato, Benjam{\'{i}}n
  S{\'{a}}nchez-Lengeling, Dennis Sheberla, Jorge Aguilera-Iparraguirre,
  Timothy~D. Hirzel, Ryan~P. Adams, and Al{\'{a}}n Aspuru-Guzik.
\newblock {Automatic Chemical Design Using a Data-Driven Continuous
  Representation of Molecules}.
\newblock {\em ACS Central Science}, 4(2):268--276, 2018.

\bibitem{he2016deep}
Kaiming He, Xiangyu Zhang, Shaoqing Ren, and Jian Sun.
\newblock Deep residual learning for image recognition.
\newblock In {\em Proceedings of the IEEE conference on computer vision and
  pattern recognition}, pages 770--778, 2016.

\bibitem{kingma2013auto}
Diederik~P Kingma and Max Welling.
\newblock Auto-encoding variational bayes.
\newblock {\em arXiv preprint arXiv:1312.6114}, 2013.

\bibitem{Dosovitskiy2020}
Alexey Dosovitskiy, Lucas Beyer, Alexander Kolesnikov, Dirk Weissenborn,
  Xiaohua Zhai, Thomas Unterthiner, Mostafa Dehghani, Matthias Minderer, Georg
  Heigold, Sylvain Gelly, Jakob Uszkoreit, and Neil Houlsby.
\newblock {An Image is Worth 16x16 Words: Transformers for Image Recognition at
  Scale}.
\newblock 2020.

\bibitem{Chen2017}
Liang-Chieh Chen, George Papandreou, Florian Schroff, and Hartwig Adam.
\newblock {Rethinking Atrous Convolution for Semantic Image Segmentation}.
\newblock 2017.

\bibitem{Belongie1998}
Serge Belongie, Chad Carson, Hayit Greenspan, and Jitendra Malik.
\newblock {Color- and texture-based image segmentation using EM and its
  application to content-based image retrieval}.
\newblock {\em Proceedings of the IEEE International Conference on Computer
  Vision}, pages 675--682, 1998.

\end{thebibliography}

\end{document}